\documentclass[preprint,12pt]{elsarticle}

\usepackage{amssymb}
\usepackage{amsmath}
\usepackage{adjustbox}
\usepackage{booktabs}
\usepackage{multirow}
\usepackage{pifont}
\usepackage{dsfont}
\newcommand{\xmark}{\ding{55}}

\usepackage{array}
\newcolumntype{L}[1]{>{\raggedright\let\newline\\\arraybackslash\hspace{0pt}}m{#1}}
\newcolumntype{C}[1]{>{\centering\let\newline\\\arraybackslash\hspace{0pt}}m{#1}}
\newcolumntype{R}[1]{>{\raggedleft\let\newline\\\arraybackslash\hspace{0pt}}m{#1}}

\usepackage{subcaption}
\usepackage{graphicx}
\usepackage{xcolor}

\makeatletter
\def\ps@pprintTitle{%
 \let\@oddhead\@empty
 \let\@evenhead\@empty
 \def\@oddfoot{\reset@font\hfil\thepage\hfil}
 \let\@evenfoot\@oddfoot
}
\makeatother

\begin{document}

\begin{frontmatter}



\title{JARViS: Detecting Actions in Video  Using Unified Actor-Scene Context Relation Modeling} 


\author[label1]{Seok Hwan Lee\fnref{label3}} 
\ead{shlee@spa.hanyang.ac.kr}

\author[label1]{Taein Son\fnref{label3}}
\ead{tison@spa.hanyang.ac.kr}

\author[label1]{Soo Won Seo}
\ead{swseo@spa.hanyang.ac.kr}

\author[label1]{Jisong Kim}
\ead{jskim@spa.hanyang.ac.kr}

\author[label2]{Jun Won Choi\corref{cor1}}
\ead{junwchoi@snu.ac.kr}

\fntext[label3]{Equal contributions}
\cortext[cor1]{Corresponding author}

\affiliation[label1]{organization={Hanyang University, Department of Electrical Engineering},
            city={Seoul},
            postcode={04763},
            country={Republic of Korea}}

\affiliation[label2]{organization={Seoul National University, Department of Electrical and Computer Engineering},
            city={Seoul},
            postcode={08826},
            country={Republic of Korea}}

\begin{abstract}
Video action detection (VAD) is a formidable vision task that involves the localization and classification of actions within the spatial and temporal dimensions of a video clip. Among the myriad VAD architectures, two-stage VAD methods utilize a pre-trained person detector to extract the region of interest features, subsequently employing these features for action detection. However, the performance of two-stage VAD methods has been limited as they depend solely on localized actor features to infer action semantics. In this study, we propose a new two-stage VAD framework called Joint Actor-scene context Relation modeling based on Visual Semantics (JARViS), which effectively consolidates cross-modal action semantics distributed globally across spatial and temporal dimensions using Transformer attention. JARViS employs a person detector to produce densely sampled actor features from a keyframe. Concurrently, it uses a video backbone to create spatio-temporal scene features from a video clip. Finally, the fine-grained interactions between actors and scenes are modeled through a Unified Action-Scene Context Transformer to directly output the final set of actions in parallel. Our experimental results demonstrate that JARViS outperforms existing methods by significant margins and achieves state-of-the-art performance on three popular VAD datasets, including AVA, UCF101-24, and JHMDB51-21.
\end{abstract}



\begin{keyword}


Deep learning \sep Action detection \sep Video action detection \sep Spatio-temporal context \sep Unified Transformer
\end{keyword}

\end{frontmatter}



\section{Introduction}
Video action detection (VAD) describes the task of simultaneously localizing action instances in both spatial and temporal domains and classifying all action instances in an untrimmed video \cite{vahdani2022deep}. VAD has recently gained considerable attention in the field of video understanding, playing an essential role in robotics, autonomous driving, video surveillance, and sports analytics. VAD is a challenging problem because it requires understanding both the spatial and temporal contexts of observed actions.
To achieve fine-grained VAD, high-level semantic information should be extracted from video frames to characterize the actions being performed.
This information can then be utilized by the VAD algorithm to accurately detect and localize specific actions within the video.

\begin{figure}[!t]
\centering
\includegraphics[width=0.85\textwidth]{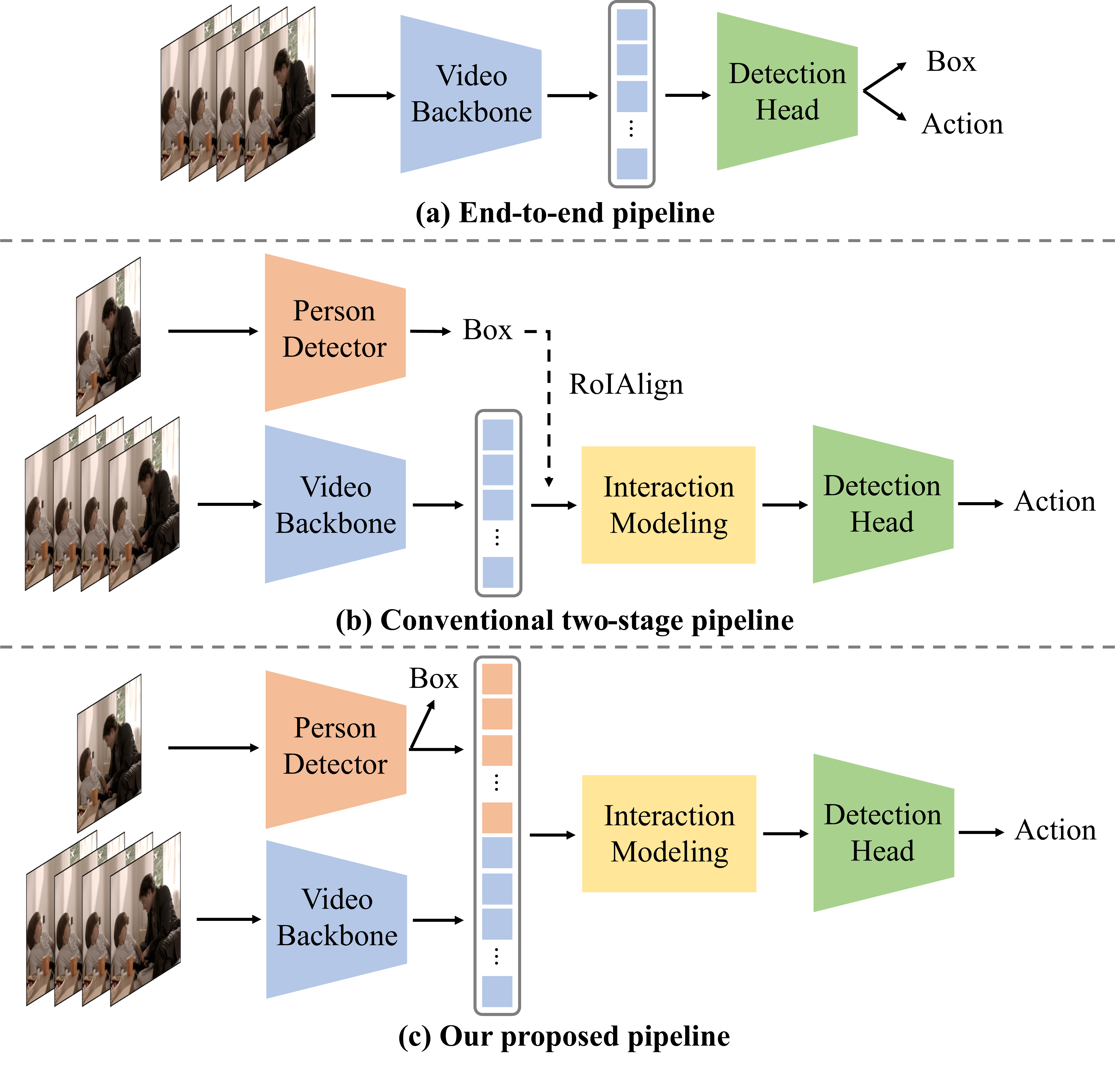}
\caption{Comparison of JARViS with other VAD architectures. JARViS employs a two-stage pipeline that initially generates densely sampled actor semantics using a pre-trained person detector, and then leverages their relationship with  spatio-temporal scene context features to produce the final set of actions.}
\label{fig:fig1}
\end{figure}

In recent years, a plethora of VAD techniques have employed a pre-trained video backbone network \cite{carreira2017quo, feichtenhofer2019slowfast, tong2022videomae} to generate spatio-temporal features.
These VAD methods can be categorized into two paradigms: end-to-end pipeline \cite{sun2018actor, chen2021watch, zhao2022tuber, wu2023stmixer} and two-stage pipeline \cite{girdhar2019video, wu2019long, pan2021actor, tong2022videomae}. Fig. \ref{fig:fig1}a and \ref{fig:fig1}b illustrate the differences between the two paradigms.
End-to-end VAD methods perform both actor detection and action classification simultaneously based on spatio-temporal features extracted from video backbone networks. Two-stage VAD methods employ a pre-trained person detector to localize actors and subsequently classify actions, utilizing the region of interest (RoI) features specific to each actor.
Lately, the end-to-end methods have garnered significant attention from researchers, resulting in rapid advancements in performance. Methods like TubeR \cite{zhao2022tuber} and STMixer \cite{wu2023stmixer} have achieved state-of-the-art (SOTA) performance by leveraging Transformer attention mechanisms to extract action-related features from the video backbone features.
In contrast, the investigation into two-stage methods has received relatively less attention recently, resulting in limited performance improvements. This is attributed to the fact that existing two-stage methods relied on local features pooled from the feature maps of the backbone for action detection, and using RoI-pooled local features could limit the modeling capacity to represent complex relational contexts present within a video.

In this study, we claim that the two-stage approach still holds potential merits and can outperform the end-to-end approach in terms of performance, provided it is meticulously designed. First, two-stage methods possess the advantage of leveraging a powerful object detection model trained on extensive object detection datasets. The cutting-edge object detection models can be seamlessly integrated into the VAD framework without any complications, and the progress in the performance of the pre-trained person detector directly contributes to enhancing the performance of the VAD model.  Moreover, the two-stage paradigm eliminates the necessity of seeking an optimal balance between actor detection and action classification tasks.
Taking these advantages into account, our study is directed towards devising a two-stage VAD architecture that effectively consolidates contextual information within videos based on actor information generated by a pre-trained person detector.

In this paper, we propose a novel two-stage VAD framework, referred to as Joint Actor-scene context Relation modeling based on Visual Semantics (JARViS), which can aggregate action-related features through a Transformer attention \cite{vaswani2017attention}. Fig. \ref{fig:fig1} highlights the difference of JARViS from the existing VAD frameworks. As shown in Fig. \ref{fig:fig1}c, JARViS first generates {\it actor proposal features} using a person detector configured to produce densely sampled actor proposals. In parallel, JARViS creates global {\it spatio-temporal scene context} using a separate 3D video backbone network such as SlowFast \cite{feichtenhofer2019slowfast} or ViT \cite{dosovitskiy2021an}. By separating backbones for actor detection and spatio-temporal scene context extraction, the resulting features are decoupled, enabling the representation of diverse action contexts crucial for VAD. Furthermore, JARViS models interactions between actor features and scene context features through {\it Unified Transformer} that has been successfully used for vision-language modeling  lately \cite{zhou2020unified}. 
By accounting for all pairwise combinations of both actor features and scene context features, the unified Transformer allows for the modeling of their fine-grained interactions, which may be crucial for the VAD task.
JARViS generates a final set of action predictions, supervised by a bipartite matching loss adapted from the end-to-end object detection framework proposed in \cite{carion2020end}. It is worth noting that since unreliable action predictions are filtered out by our detection head at the final stage, JARViS can leverage densely sampled actor proposals obtained from the person detector for joint actor-scene context relation modeling.

This approach differs from the traditional frameworks which are illustrated in Fig. \ref{fig:fig1}a and Fig. \ref{fig:fig1}b. In these conventional paradigms, the VAD task is either executed end-to-end directly from the video clip \cite{sun2018actor, girdhar2019video, li2020actions, chen2021watch, zhao2022tuber, wu2023stmixer}, or the output of the person detector is merely utilized to find Regions of Interest (RoI) boxes, which are then used to extract features from the video backbone \cite{gu2018ava, zhang2019structured, tang2020asynchronous, pan2021actor}. The first approach, the end-to-end architecture, is restricted in its capacity to model interactions between actors and their surroundings. The second approach limits the role of the person detector to merely identifying areas of interest for VAD, rather than contributing to feature construction. In contrast, our JARViS efficiently captures the intricate relationship between densely sampled actor features and global scene context features, enabling fine-grained action detection.
To our best  knowledge, we are the first to propose a two-stage VAD architecture that aggregates action contexts on a global scale to produce action semantics.

Furthermore, we extend JARViS to a VAD task with long-term video clips. We employ a weighted score aggregation method that combines action classification scores obtained between a single keyframe and multiple short video clips to gather long-term context information. This approach significantly enhances accuracy and simplifies the training process compared to conventional long-term VAD methods that rely on feature memory banks \cite{wu2019long, tang2020asynchronous, pan2021actor} or long-term memory encoding \cite{zhao2022tuber}.

Surprisingly, this straightforward JARViS structure surpassed existing SOTA VAD methods, TubeR \cite{zhao2022tuber}, and STMixer \cite{wu2023stmixer}.

\begin{itemize}
\item We propose a novel two-stage VAD architecture called JARViS. As depicted in Fig. \ref{fig:fig1}, JARViS is different from traditional two-stage VAD methods. Rather than relying solely on local actor Region of Interest (RoI) features for extracting action semantics, JARViS separately encodes both actor contexts and spatio-temporal scene contexts and integrates these contexts, employing a unified Transformer. The unified Transformer effectively models actor-to-actor and actor-to-scene interactions through its self-attention.

\item We introduce a new long-term VAD method based on weighted score aggregation. This approach simplifies the learning process significantly compared to existing long-term VAD methods by eliminating the need for long-term feature banks.

\item The proposed JARViS achieves SOTA performance on popular VAD benchmarks and outperforms the latest end-to-end VAD methods by significant margins.
\end{itemize}

\section{Related Work}

\subsection{End-to-end VAD Methods}
End-to-end VAD methods concurrently optimize the entire network for both the action localization and classification tasks. WOO \cite{chen2021watch} integrated both actor detection head and action classification head with 3D CNN-based backbone features. ACRN \cite{sun2018actor} and VTr \cite{girdhar2019video} encoded pairwise relations from cropped actor features and a global feature map using actor-centric relation networks.
MOC \cite{li2020actions} introduced an action tubelet detection framework that predicts action instances from trajectories of moving points.
Recently, TubeR \cite{zhao2022tuber} and STMixer \cite{wu2023stmixer} have employed Transformer attention to gather multi-scale spatio-temporal features as scene context. EVAD \cite{chen2023efficient} designed a keyframe-centric token pruning module to remove tokens irrelevant to actor motions and scene context. STAR \cite{gritsenko2024end} introduced a versatile model that leverages entire tubelet annotations or sparse bounding boxes on individual frames.

\subsection{Two-stage VAD Methods}
Two-stage VAD methods apply an off-the-shelf person detector to a keyframe of a video clip to localize actors, and  then detect action instances using the extracted actor information.
In \cite{zhang2019structured}, an actor-centric graph was constructed based on a set of tubes generated for the objects tracked by an object tracker to capture human-human or human-object interactions.
AIA \cite{tang2020asynchronous} modeled person-person, person-object, and temporal interactions between the extracted object features.
Similarly, ACAR \cite{pan2021actor} utilized Transformer attention mechanisms to model higher-order pairwise relationships between any two actors. Recently, various methods \cite{tong2022videomae, wang2023videomae, wang2023masked, ryali2023hiera} have adopted vision transformers pre-trained on powerful visual pretext tasks, such as MAE \cite{he2022masked}, to achieve superior performance.

The aforementioned methods  adopted an actor-centric approach, inferring action semantics from actor features extracted through RoIAlign \cite{he2017mask} from video backbone features. On the contrary, JARViS deduces action semantics in a non-centric manner using densely sampled actor features along with the overarching global scene context features.


\begin{figure*}[!t]
\centering
\includegraphics[width=0.95\textwidth]{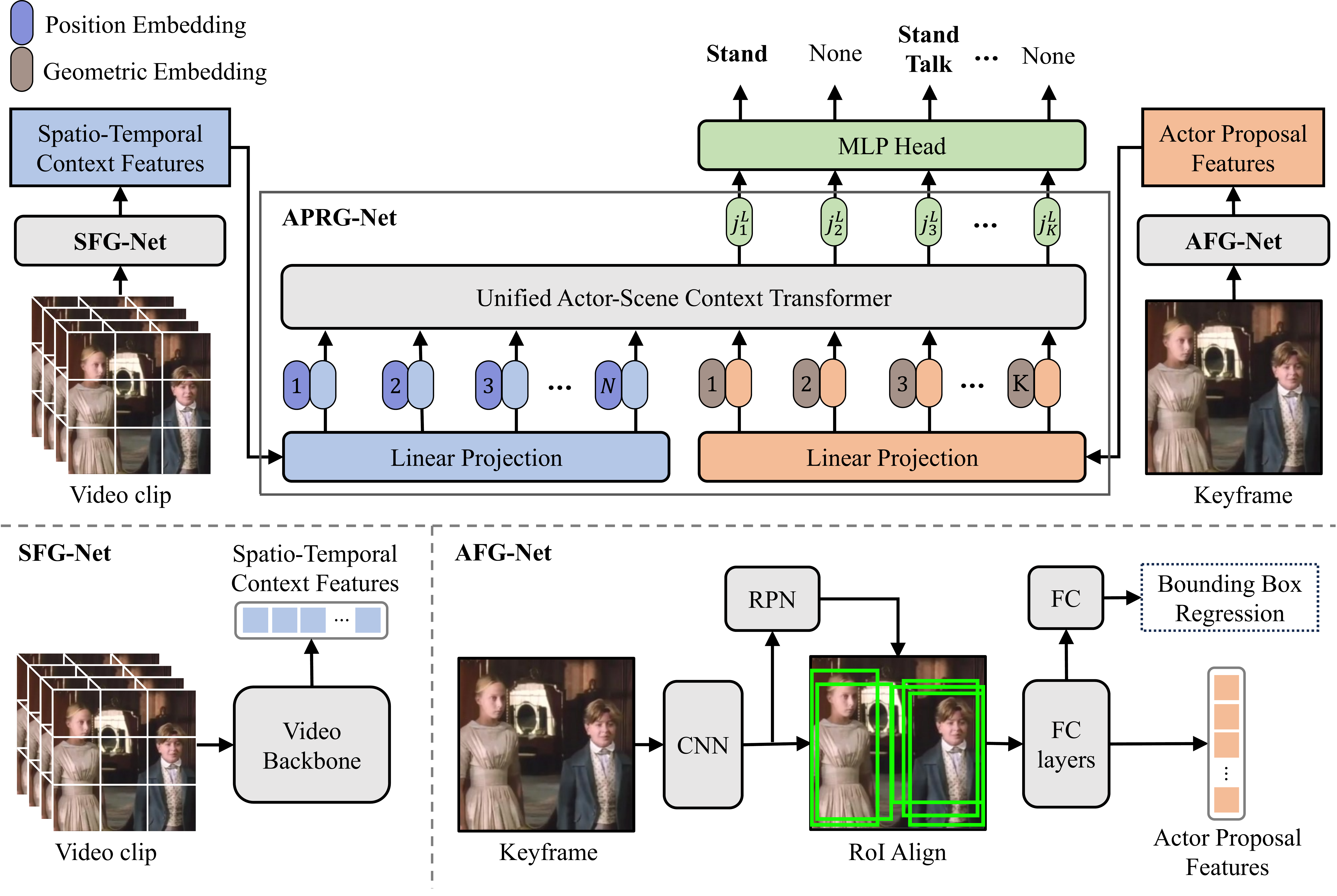}
\caption{Overall architecture of JARViS model. JARViS produces actor proposal features and scene context features by applying separate backbone networks to the keyframe image and the video clip, respectively. These features are linearly mapped to the embedding vectors of the same size. The transformer encoder then transforms the embedding vectors into the final action proposal features. Finally, the action classification results are obtained through the MLP head.
}
\label{fig:fig2}
\end{figure*}

\section{Methodology}

\subsection{Overview}
The overall structure of JARViS is depicted in Fig. \ref{fig:fig2}.
JARViS consists of two primary components: the Semantic Feature Extraction Network (SE-Net) and the Action Proposal Generation Network (APRG-Net). The SE-Net has two parallel branches: Actor Feature Generation Network (AFG-Net) and Spatio-Temporal Feature Generation Network (SFG-Net). AFG-Net produces the actor proposal features from a target keyframe, and SFG-Net generates the spatio-temporal scene context features from a sequential video clip. These two branch networks offer a set of independently generated  embedding features for VAD.  APRG-Net deduces action-related semantics by modeling pairwise interactions among the multi-modal features.   Finally, APRG-Net produces the action proposal features for each actor candidate and predicts a set of actions using the multi-layer perceptron (MLP) head.

\subsection{Semantic Feature Extraction Network (SE-Net)}
SE-Net generates two action-related embedding features: actor features and spatio-temporal scene features using AFG-Net and SFG-Net, respectively.
\subsubsection{Actor feature generation network (AFG-Net)}
Given an input keyframe image $I_t$ at a given time step $t$, we extract $K$ Region of Interest (RoI) features using an off-the-shelf person detector such as Faster R-CNN \cite{ren2015faster} or DETR \cite{carion2020end}.  These features, denoted as $f_a \in \mathbb{R}^{C\times K}$, are obtained from the CNN feature maps using RoIAlign \cite{he2017mask}, where $C$ represents the channel dimension of the actor features. By applying a low confidence score threshold with the person detector, AFG-Net generates densely populated actor proposal features, which offer diverse contextual information on actors.
The $K$ actor features are fed into two fully-connected layers to predict the actors' confidence scores $\mathbf{\hat{h}} = \{\hat{h}_i\}_{i=1}^{K}$ and bounding boxes $\mathbf{\hat{b}} = \{\hat{b}_i\}_{i=1}^{K}=\{(x_i^{lt},y_i^{lt},x_i^{rb},y_i^{rb})\}_{i=1}^{K}$, where $(x^{lt},y^{lt})$ and $(x^{rb},y^{rb})$ denote the normalized coordinates of the top-left and bottom-right corners of 2D boxes, respectively.
These actor proposal features are then interacted with spatio-temporal scene features to facilitate action detection in the subsequent APRG-Net.

\subsubsection{Spatio-temporal feature generation network (SFG-Net)}
Given an input video clip $I_{t-T_p:t+T_f}=\{ I_{t-T_p}, ..., I_{t+T_f}\}$ of $T=T_p+T_f+1$ frames, SFG-Net extracts the spatio-temporal feature map of size $H\times W \times T$ using the video backbone network such as SlowFast \cite{feichtenhofer2019slowfast} and ViT \cite{dosovitskiy2021an}.
The spatio-temporal feature map is then flattened to the scene context features $f_v \in \mathbb{R}^{C' \times N }$, where $C'$ is their channel size and $N=H\times W \times T$ is the number of the flattened features. SFG-Net can be pre-trained on the action recognition dataset such as Kinetics-400 \cite{kay2017kinetics} and Kinetics-700 \cite{carreira2019short}.



\subsection{Action Proposal Generation Network (APRG-Net)}
APRG-Net infers the relationship between the actor-centric and spatio-temporal scene features  using  {\it Unified Actor-Scene Context Transformer}. The Unified Actor-Scene Context Transformer employs $L$ layers of multi-head transformer encoding to generate action proposal features.
APRG-Net first linearly maps the actor features into the embedding vectors of dimension $D$ as
\begin{align}
    a= E_a f_a  + E_g g ,
\end{align}
where $E_a \in \mathbb{R}^{D \times C}$ and $E_g \in \mathbb{R}^{D \times 6}$ are the trainable weights for linear projection, and $g \in \mathbb{R}^{6 \times K}$ denotes the geometric information of $K$ actor proposals characterized by a six-dimensional vector $[x^{lt},y^{lt},x^{rb},y^{rb}, w, h]^\top$, with $w$ and $h$ denoting the width and height of the associated bounding box, respectively.
In parallel, the embedding vectors for the scene context features are obtained from
\begin{align}
   v = E_v f_v + E_{pos},
\end{align}
where $E_v \in \mathbb{R}^{ D \times C'}$ is the  weights of the projector, and $E_{pos} \in \mathbb{R}^{ D \times N}$ is the sinusoidal positional encoding \cite{vaswani2017attention}. In our setup, the size $D$ of the embedding vectors was set to $256$. We construct the input to the transformer encoder as $j^0=[ a,  v] = [a_1 ..., a_K, v_1, ..., v_N] \in \mathbb{R}^{D \times (K+N)}$. The input embeddings are then passed through the $L$ layers of transformer encoding  as
\begin{align}
    &\qquad \qquad z^l = \text{MSA}(\text{LN}(j^{l-1})) + j^{l-1},\\
    &\qquad \qquad j^l = \text{MLP}(\text{LN}(z^{l})) + z^{l},   \quad  \quad l=1, ..., L,
\end{align}
where MSA denotes multi-head self-attention \cite{vaswani2017attention},  LN denotes layer normalization \cite{ba2016layer}, and MLP denotes a multilayer perceptron operation.
The first $K$ embeddings $[j_1^{L},...,j_K^{L}]$ of  $j^L=[j_1^{L}, ..., j_{(K+N)}^{L}]$ correspond to the features associated with densely sampled $K$ actor proposals. Then, these $K$ embeddings are fed into the classification head to predict a set of action class scores $[\hat{c}_1,...,\hat{c}_K] \in \mathbb{R}^{N_{cls} \times K}$.
The set of the action class scores and the actor bounding boxes constitutes a set of action instances $(\mathbf{\hat{b}}, \mathbf{\hat{c}})=\{(\hat{b}_{i},\hat{c}_{i})\}_{i=1}^{K}$. Finally, the set of final $K' (< K)$ action instances is selected based on their confidence scores. Note that even if JARViS starts with densely sampled actor proposals of size $K$, it rejects $(K-K')$ unreliable action instances based on the relational reasoning.

\begin{figure*}[t]
\centering
\begin{subfigure}[b]{0.3\textwidth}
\centering
\includegraphics[width=\textwidth]{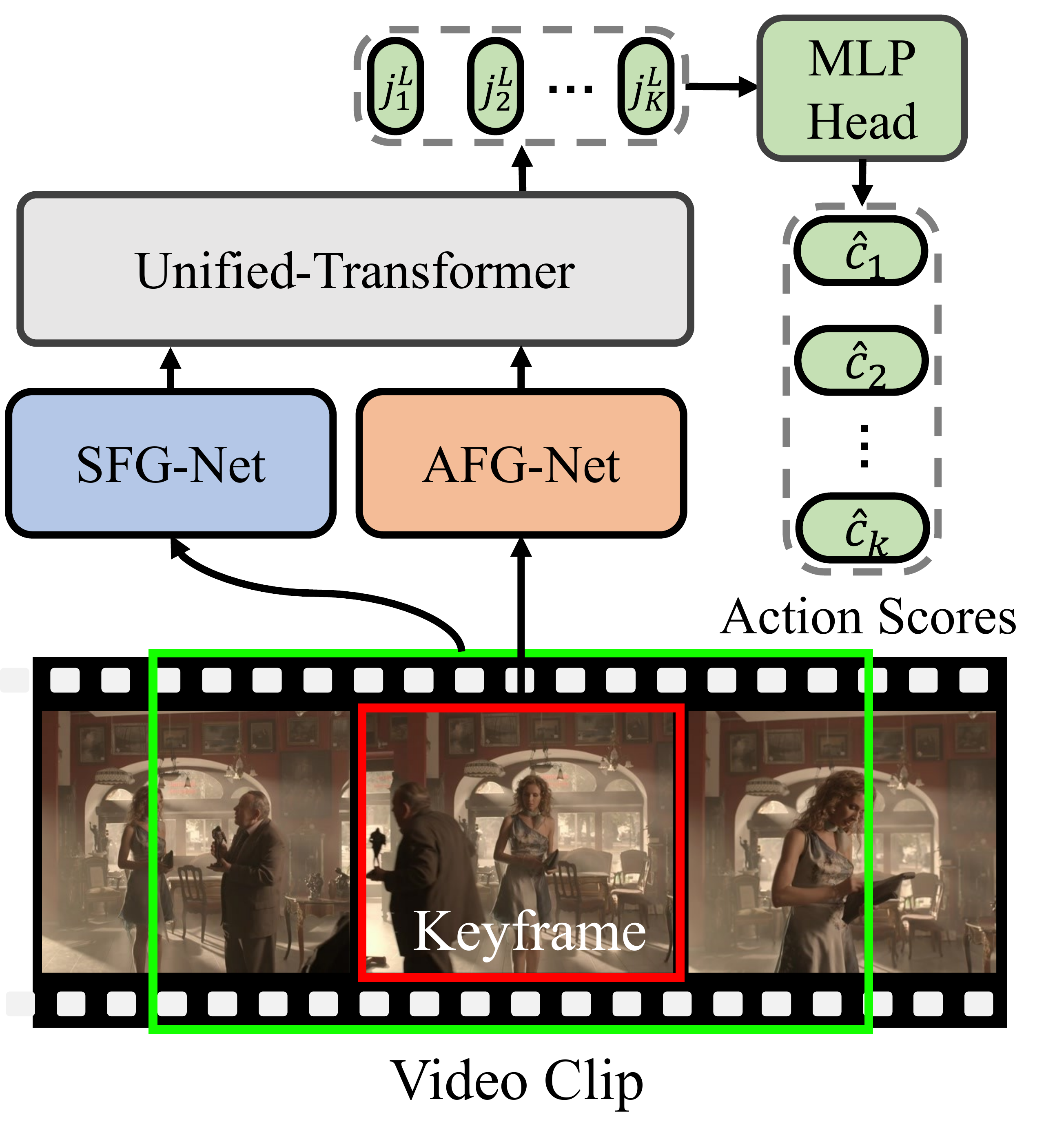}
\caption{JARViS for short-term VAD}
\label{fig:fig3b}
\end{subfigure}
\hfill
\begin{subfigure}[b]{0.66\textwidth}
\centering
\includegraphics[width=\textwidth]{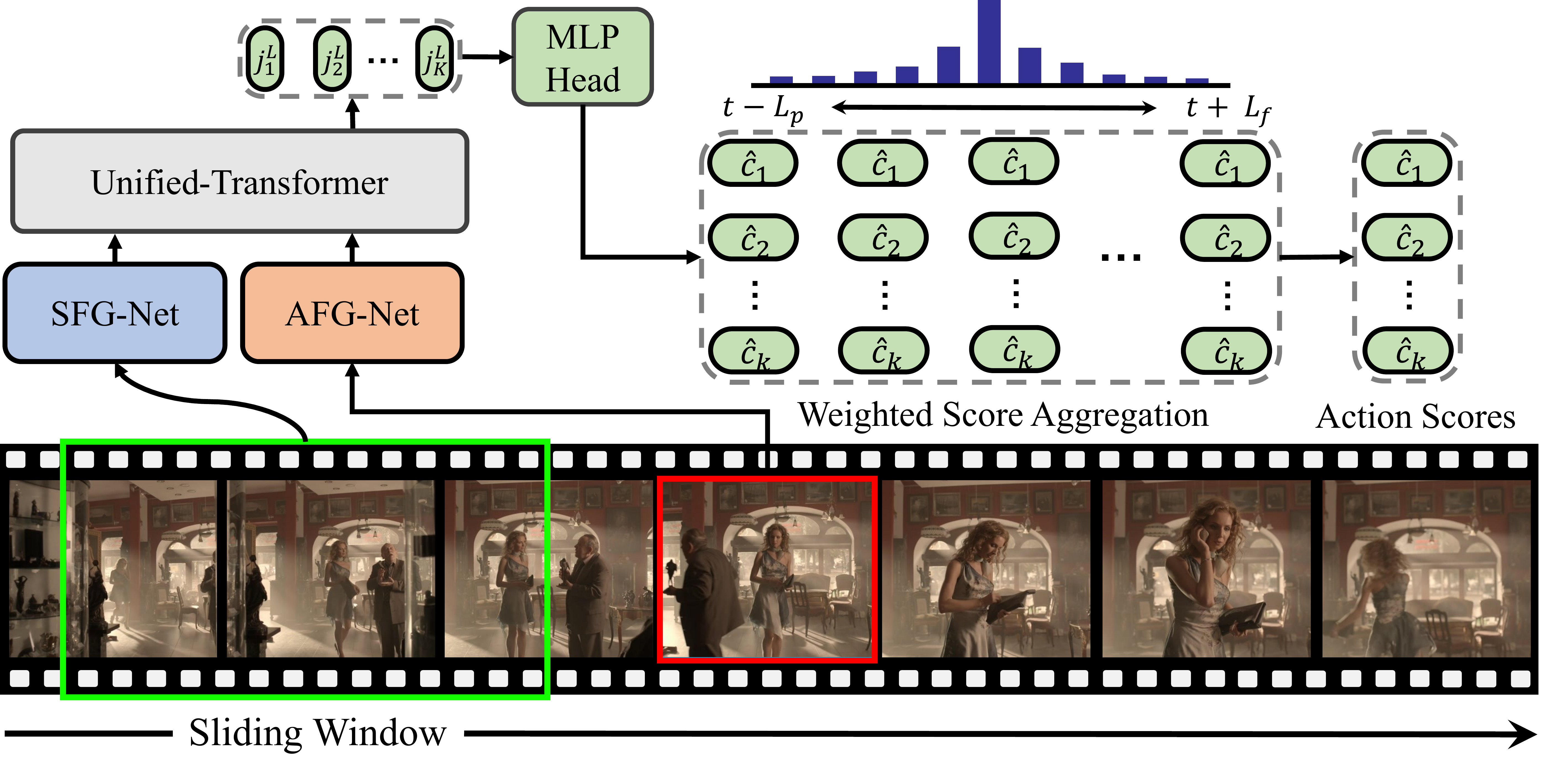}
\caption{JARViS for long-term VAD}
\label{fig:fig3c}
\end{subfigure}
\caption{\textbf{JARViS with a long-term video clip. } JARViS can detect actions from a long-term video clip. Given a fixed keyframe, JARViS produces the action scores based on a short-term video sequence in a sliding window. These action scores, obtained each time the window moves, are aggregated with trainable weights. Note that the combined weights vary depending on the action class and the relative window position away from the keyframe.}
\label{fig:fig3}
\end{figure*}

\subsection{Weighted Score Aggregation for Long-Term VAD}
We can also extend JARViS for the task of performing long-term VAD that deals with longer video clips. In this task, actions should be detected based on both a keyframe $I_t$ and long-length video frames $\{I_{t-L_p}, \cdots, I_{t+L_f}\}$ ($L_f+L_p+1 \gg T$) around the keyframe.
The key idea is to aggregate the action information obtained by applying JARViS at different time stamps within a long video clip. (See the illustration in Fig. \ref{fig:fig3}.) To this end, a sliding window is applied to $\{I_{t-L_p}, \cdots, I_{t+L_f}\}$ with a stride $W$ to retrieve short-term video clips of $T$ frames, $I(n)=\{I_{t+Wn-T_p},...,I_{t+Wn+T_f}\}$, where $n \in [-\lfloor L_p/W \rfloor , \lfloor L_f/W \rfloor ]$. The output of JARViS with the input $I(n)$ is obtained as
\begin{align}
    (\mathbf{\hat{b}}_n, \mathbf{\hat{c}}_n) = {\rm JARViS}(I(n), I_t),
\end{align}
where ${\rm JARViS}(I(n), I_t) $ implies the operation of JARViS taking the keyframe $I_t$ and the short-term video clip $I(n)$ as inputs.  Note that JARViS is capable of detecting actions even when the key frame $I_t$ is not within the short frames $I(n)$ owing to our {\it Temporal Data Augmentation Strategy}, which is described in {\it Implementation Details} Section. Based on the $K$ actor proposals obtained from the keyframe $I_t$, the proposed {\it weighted score fusion method} aggregates the action scores across window position $n$, i.e.,
\begin{align}
\hat{c}_{k}^{long-term} = \sum_{n=-\lfloor L_p/W \rfloor}^{ \lfloor L_f/W \rfloor}A_{n,k} \hat{c}_{n,k},
\end{align}
where $\hat{c}_{n,k}$ and $A_{n,k}$ are the score for the $k$th action class and the $n$th window position in $\mathbf{\hat{c}}_n$ and the corresponding combining weight, respectively. Note that aggregation weights vary depending on a relative window position from the keyframe and an action class.
Our evaluation shows that our method promises better long-term VAD performances than the existing long-term VAD methods \cite{wu2019long, tang2020asynchronous, pan2021actor, zhao2022tuber}.

\subsection{Loss Function}

JARViS processes a set of $K$ action instances  obtained by the model, where $K$ is set to be larger than the typical number of actors in a keyframe. We associate these action instances with the ground truth action instances via bipartite matching. We adopt a training procedure similar to that of DETR \cite{carion2020end}.



\subsubsection{Bipartite matching}

The JARViS model matches the set of $K$ predictions $\{(\hat{b}_i, \hat{h}_i, \hat{c}_i)\}_{i=1}^{K}$ with
the set of $K^{gt}$ ground truth actions $\{({b}_i^{gt}, {h}_i^{gt}, {c}_i^{gt})\}_{i=1}^{K^{gt}}$ that exist in the keyframe, where $\hat{b}_i$, $\hat{h}_i$, and $\hat{c}_i$ are the predicted bounding box, person detection confidence score, and action classification score for the $i$th action instance, respectively, and
 ${b}_i^{gt}$, ${h}_i^{gt}$ and ${c}_i^{gt}$ are the ground truth bounding box, the target label for binary classification, and the target label for action classification, respectively. We construct the set of ground truth actor targets $\mathbf{y}^{gt}=\{y_i^{gt}\}_{i=1}^{K}$ as
\begin{align}
y_i^{gt} = \left\{
\begin{matrix}
(b_i^{gt},h_i^{gt}) & \quad \text{if} \quad 1 \leq i \leq K^{gt} \\
(\varnothing,\varnothing), & \quad \text{otherwise.}
\end{matrix}
\right.
\end{align}
Then, we match the set of predicted output $\mathbf{\hat{y}}=\{\hat{y}_i\}_{i=1}^{K}=\{(\hat{b}_i, \hat{h}_i)\}_{i=1}^{K}$ with the target $\mathbf{y}^{gt}$ using the bipartite matching.
Following Deformable DETR \cite{zhu2021deformable}, we use the matching cost $\mathcal{L}_{\text{match}}(y_i^{gt}, \hat{y}_{j})$
\begin{align}
\mathcal{L}_{\text{match}}(y_i^{gt}, \hat{y}_{j}) = \mathds{1}_{\{h_i^{gt}\neq\varnothing\}} \mathcal{L}_{\text{cls}}(h_i^{gt},\hat{h}_{j}) + \mathds{1}_{\{h_i^{gt}\neq\varnothing\}} \mathcal{L}_{\text{box}}(b_i^{gt},\hat{b}_{j}),
\end{align}
where $\mathcal{L}_{\text{cls}}$ denotes the sigmoid focal loss \cite{lin2017focal} and $\mathcal{L}_{\text{box}}$ denotes a linear combination of $l1$ loss and GIoU loss \cite{rezatofighi2019generalized}. Note that the term $\mathcal{L}_{\text{cls}}(h_i^{gt},\hat{h}_{j})$ is affected by the person detection confidence score, not the action classification score. Our empirical findings indicate that incorporating the action classification score into the matching loss does not enhance the overall performance. This will be confirmed in Section \ref{sec:matching_cost}.

Finally, we use Hungarian algorithm \cite{kuhn1955hungarian} to find the optimal bipartite matching between $ \mathbf{\hat{y}} $ and $\mathbf{y}^{gt}$
\begin{align}
\hat{\sigma} &= \operatorname*{argmin}_{\sigma \in \mathfrak{S}_K}\sum_{i=1}^{K}\mathcal{L}_{\text{match}}(y_i^{gt}, \hat{y}^p_{\sigma(i)}),
\end{align}
where $\sigma$ represents an arrangement of $K$ indices obtained by Hungarian algorithm, and $\sigma(i)$ denotes the $i$th element of $\sigma$.
The optimal matching is then used to obtain the loss function for training the JARViS.

\subsubsection{Loss function for JARViS}
After the result $\sigma$ of bipartite matching  is obtained, we optimize the entire JARViS network using the action classification loss
\begin{align} \label{eq:loss_cls}
\mathcal{L}(\mathbf{c}^{gt},\mathbf{\hat{c}}) = \sum_{i=1}^K \mathcal{L}_{\text{cls}}(c_i^{gt},\hat{c}_{{\sigma}(i)}),
\end{align}
where $\mathbf{\hat{c}} = \{\hat{c}_i\}_{i=1}^K$,   $\mathbf{c}^{gt}=\{g_i^{gt}\}_{i=1}^{K}$, and
\begin{align}
g_i^{gt} = \left\{
\begin{matrix}
c_i^{gt} & \quad \text{if} \quad 1 \leq i \leq K^{gt} \\
\varnothing, & \quad \text{otherwise.}
\end{matrix}
\right.
\end{align}

For a long-term VAD, we train JARViS model with short-term video clips and then optimize the score aggregation weights using long-term video clips while freezing the trained JARViS model.  Note that we optimize the score aggregation weights using the set of the long-term action scores $\mathbf{\hat{c}}^{long-term}$ instead of $\mathbf{\hat{c}}$ in (\ref{eq:loss_cls}).

\section{Experiments}
\subsection{Datasets \& Metrics}
    We evaluate the proposed JARViS on three mostly used VAD datasets \textbf{AVA} \cite{gu2018ava}, \textbf{UCF101-24} \cite{soomro2012ucf101}, and \textbf{JHMDB51-21} \cite{jhuang2013towards}. The AVA dataset comprises  movie clips of 15-minute duration, with a training/validation split of 235/64. We assess our approach on 60 action classes within AVA v2.1 and AVA v2.2. UCF101-24 is a VAD dataset derived from 24 sports-related classes in the UCF101 dataset of realistic action videos. With a total of 3,207 instances, we conduct evaluations on the first split, in line with prior studies. JHMDB51-21 is a subset of the JHMDB51 dataset, containing 928 trimmed video clips across 21 classes. We employed a standard protocol involving three training/test splits and used the average performance of these three test splits. We reported the frame level mean Average Precision (mAP) at IoU threshold of 0.5 for all three datasets. We also compare the video level mAP at IoU threshold of 0.2 and 0.5 on JHMDB51-21.

\subsection{Implementation Details}\label{Implementation details}

\subsubsection{Configurations} We used video clips of 2.1 seconds to train JARViS. For long-term VAD task, we used video clips of 12 seconds. The JARViS trained for a long-term VAD task is called JARViS-LT. The parameters $T_p$ and $T_f$ are set to the same value, and  $L_p$ and $L_f$ are also set to the same value.  When configuring JARViS, we followed the hyperparameter settings of the original backbone networks. The configuration of the {\it Unified Action-Context Transformer} is provided in Table \ref{arc_details}.

\begin{table}[!h]
\caption{Architecture details of unified Transformer.}
\label{arc_details}
\centering

\begin{tabular}{l|cc}

\toprule

Hyperparameters           & Unified Transformer   \\

\midrule

Number of Layers            &     6             \\
Normalized before           &    yes              \\
Hidden size                 &    256                 \\
FFN hidden size             &   1024              \\
Attention heads             &   8               \\
Activation function         &  GELU           \\
Dropout rate                &    0.1             \\
Attention dropout rate      &    0.1             \\

\bottomrule

\end{tabular}
\end{table}

\subsubsection{Person detector}
    For AFG-Net, we employed either a Faster R-CNN \cite{ren2015faster} with a ResNeXt-101-FPN \cite{lin2017feature, xie2017aggregated} backbone trained with Detectron \cite{girshick2018detectron} or DETR \cite{carion2020end} with a ResNet-50 \cite{he2016deep}. The AFG-Net was pre-trained on ImageNet \cite{russakovsky2015imagenet} and COCO human keypoint images \cite{{lin2014microsoft}}, and then fine-tuned for each of the target VAD datasets.

\subsubsection{Video backbone networks}
     For SFG-Net, we have used both 3D CNN-based backbones and ViT-based backbones as a video backbone network. SlowFast-R50 and SlowFast-R101 were used as a 3D-CNN-based backbone, where were pre-trained on Kinetics-400 (K400) \cite{kay2017kinetics} and Kinetics-700 (K700) dataset \cite{carreira2019short}, respectively. We configured SlowFast backbone following the suggestion of \cite{feichtenhofer2019slowfast}. ViT-B was used as a ViT-based backbone. For a fair comparison, we initialized ViT-B backbone network with the pre-trained weights of VideoMAE \cite{tong2022videomae} or VideoMAE v2 \cite{wang2023videomae}.


\subsubsection{Training and inference}
    While the JARViS model was trained, the pre-trained person detector was frozen. For JARViS-LT,  we first trained JARViS with the short-term video clips and fine-tuned the score aggregation weights for the long-term VAD task while keeping the JARViS model frozen.
    Our models were trained using the AdamW optimizer \cite{loshchilov2017decoupled} with a 1e-4 weight decay.
    The initial learning rates were set to 1e-5 for the video backbone and 1e-4 for all other networks and then decayed by a factor of $10$ at the 6th epoch. We trained the model for 8 epochs with a batch size of 16.
    For AFG-Net, we used the original input image without changing its resolution. For SFG-Net, we scaled the shorter spatial side of the input video clip to 256 pixels for the SlowFast backbone and 224 pixels for the ViT and Hiera backbone, while maintaining the aspect ratio.

    We used four NVIDIA GeForce RTX 3090 GPUs to train JARViS model except when Hiera-L backbone was used on AVA dataset \citep{gu2018ava}, UCF101-24 \citep{soomro2012ucf101} and JHMDB51-21 \citep{jhuang2013towards}. We used eight A100 GPUs when Hiera-L backbone was used.

\subsubsection{Data augmentation}
    We tried random horizontal flipping for data augmentation. We also devised a {\it temporal data augmentation} for JARViS. Given a keyframe image $I_t$, the time offset $\Delta$ of the input video clip $\{I_{t-T_p+ \Delta},...,I_{t+T_f+ \Delta}\}$ was randomized during training. The range of time offset was set to $\pm 1.5$ seconds with respect to the position of the keyframe. This data augmentation allows the model to utilize the spatio-temporal information within the video clip appropriately, even when the video clip is not temporally aligned with the keyframe.


\begin{table*}[t]

\centering
\caption{Performance evaluation on AVA dataset. The input sizes are shown as the frame number and corresponding sample rate. Duration refers to the total video length in seconds, accounting for the model's incorporation of long-term context. ViT-B marked with $\text{}^{\ast}$ is initialized with the pre-trained weights on VideoMAE v2.}
\label{table_ava_v2.2_overall}
\begin{adjustbox}{width=0.99\textwidth}
\begin{tabular}{lccccccc}


\toprule

\multicolumn{1}{c}{\multirow{2}{*}{\textbf{Model}}} & \multirow{2}{*}{\textbf{Detector}} & \multirow{2}{*}{\textbf{Input}} & \multirow{2}{*}{\textbf{Backbone}} & \multirow{2}{*}{\textbf{Pre-train}} & \multirow{2}{*}{\textbf{Duration}} & \multicolumn{2}{c}{\textbf{mAP}} \\
\cmidrule(r){7-8}
\multicolumn{1}{c}{}                                &                                    &                                                               &                              & &                                        & v2.1     & v2.2 \\

\midrule
\multicolumn{4}{l}{\textbf{Compared to two-stage VAD methods with a person detector} }                             & \quad          &  \quad & \quad  & \\
\midrule
LFB \cite{wu2019long}                        & F-RCNN          & 32 $\times$ 2         & I3D-R101-NL       & K400            & 60        & 27.7  &  -            \\
SlowFast  \cite{feichtenhofer2019slowfast}   & F-RCNN          & 32 $\times$ 2         & SF-R101-NL         & K600           & 2         & 28.2  & 29.0 \\
AIA  \cite{tang2020asynchronous}             & F-RCNN          & 32 $\times$ 2         & SF-R101            & K700           & 62        & 31.2  & 32.3 \\
ACAR  \cite{pan2021actor}                    & F-RCNN          & 32 $\times$ 2         & SF-R101            & K400           & 21        & 30.0  &  - \\
ACAR  \cite{pan2021actor}                    & F-RCNN          & 32 $\times$ 2         & SF-R101            & K700           & 21        &  -    & 33.3 \\
VideoMAE  \cite{tong2022videomae}            & F-RCNN          & 16 $\times$ 4         & ViT-B              & K400           & 2         & - & 31.8 \\
MVD \cite{wang2023masked}      & F-RCNN       & 16 $\times$ 4         & ViT-B     & IN-1K + K400    & 2         & - & 34.2 \\

\midrule

\textbf{JARViS}                              & F-RCNN          & 32 $\times$ 2         & SF-R101           & K700          & 2      &  34.0 & 35.1 \\
\textbf{JARViS-LT}                           & F-RCNN          & 32 $\times$ 2         & SF-R101           & K700          & 12     &  \textbf{34.9}  & \textbf{36.3} \\
\textbf{JARViS}                              & F-RCNN          & 16 $\times$ 4         & ViT-B             & K400          & 2      & - & 35.4     \\
\midrule
\multicolumn{4}{l}{\textbf{Compared to end-to-end VAD methods} }                               &  \quad            & \quad      & \quad    &\\
\midrule
ACRN \cite{sun2018actor}                     & \xmark          & 20 $\times$ 1         &  S3D-G            & K400           & 1        & 17.4  &  -            \\
VTr \cite{girdhar2019video}                  & \xmark          & 64 $\times$ 1         & I3D-VGG           & K400           & 2        & 24.9  &  -            \\
WOO  \cite{chen2021watch}                    & \xmark          & 32 $\times$ 2         & SF-R101-NL        & K600           & 2        & 28.0  & 28.3 \\
TubeR  \cite{zhao2022tuber}                  & \xmark          & 32 $\times$ 2         & CSN-152           & IG-65M + K400      & 15       & 32.0  & 33.6 \\
STMixer  \cite{wu2023stmixer}               & \xmark          & 32 $\times$ 2         & SF-R101           & K700        & 62       & 32.6  & 32.9 \\
STMixer  \cite{wu2023stmixer}               & \xmark          & 16 $\times$ 4         & $\text{ViT-B}^{\ast}$         & K710 + K400    & 2        & -     & 36.1 \\
EVAD  \cite{chen2023efficient}                 & \xmark          & 16 $\times$ 4         & $\text{ViT-B}^{\ast}$         & K710 + K400    & 2        & -     & 37.7 \\
\midrule

\textbf{JARViS}                             & DETR             & 32 $\times$ 2         & SF-R101                           & K700           & 2        & 34.0           & 34.9  \\
\textbf{JARViS-LT}                          & DETR             & 32 $\times$ 2         & SF-R101                           & K700           & 12       & \textbf{35.0}  & 35.9  \\
\textbf{JARViS}                             & DETR             & 16 $\times$ 4         & $\text{ViT-B}^{\ast}$            & K710 + K400    & 2        &  -             & 39.5 \\
\textbf{JARViS-LT}                          & DETR             & 16 $\times$ 4         & $\text{ViT-B}^{\ast}$            & K710 + K400    & 12       &  -             & \textbf{40.0} \\

\bottomrule

\end{tabular}
\end{adjustbox}

\end{table*}

\subsection{Main Results}

\begin{table}[t]
\caption{Performance comparison on UCF101-24 dataset with frame-mAP@0.5.}
\label{table_ucf}
\centering
\begin{adjustbox}{width=0.6\linewidth}
    \centering
    \begin{tabular}{lccccc}
        \toprule
        Model                           &Input               & Backbone  & Pre-train  & mAP \\
        \midrule
        ACT \cite{kalogeiton2017action} &\, 6 $\times$ 1     & VGG  & ImageNet & 67.1 \\
        STEP \cite{yang2019step} &\, 6 $\times$ 1     & VGG  & ImageNet & 75.0 \\
        MOC \cite{li2020actions} &\, 7 $\times$ 1     & DLA34& COCO & 78.0 \\
        AIA \cite{tang2020asynchronous}&32 $\times$ 1       & C2D  & K400 & 78.8 \\
        ACAR \cite{pan2021actor} &32 $\times$ 1   & SF-R50  & K400 & 84.3 \\
        TubeR \cite{zhao2022tuber} &32 $\times$ 2       & CSN-152 & K400 & 83.2 \\
        STMixer \cite{wu2023stmixer} &32 $\times$ 2   & SF-R101-NL  & K600 & 83.7 \\
        \midrule
        \textbf{JARViS} &32 $\times$ 2   & SF-R50  &K400 & \textbf{84.6} \\
        \bottomrule
    \end{tabular}
\end{adjustbox}
\end{table}

\begin{table}[t]
\caption{Performance comparison on JHMDB51-21 dataset with video-mAP. f-mAP denotes the frame mAP@0.5.}
\label{table_jhmdb}
\centering
\begin{adjustbox}{width=0.6\linewidth}
    \centering
    \begin{tabular}{lccccc}
        \toprule
        Model                           &Input & Backbone & f-mAP & 0.20 & 0.50\\
        \midrule
        ACT  &\, 6 $\times$ 1   & VGG   & 65.7 & 74.2& 73.7 \\
        MOC &\, 7 $\times$ 1   & DLA34  & 70.8 & 77.3& 77.2\\
        AVA &20 $\times$ 1     & I3D   & 73.3 & -& 78.6\\
        ACRN &20 $\times$ 1     & S3D-G  &  77.9 & -& 80.1\\
        WOO &32 $\times$ 2     & SF-R101-NL &  80.5 & -& -\\
        TubeR &32 $\times$ 2     & CSN-152 &  - & 87.4& 82.3\\
        STMixer &32 $\times$ 2     & SF-R101-NL & 86.7 & -& -\\
        \midrule
        \textbf{JARViS} &32 $\times$ 2     & SF-R101  &\textbf{89.1} &\textbf{91.4} &\textbf{89.2}\\
        \bottomrule
    \end{tabular}
\end{adjustbox}
\end{table}

\subsubsection{Results on AVA dataset}
    Table \ref{table_ava_v2.2_overall} provides the performance of JARViS in comparison with the latest VAD methods on the validation set of AVA v2.1 and AVA v2.2.  We consider the VAD methods with 3D-CNN backbones and those with ViT backbones separately for comparison. Recall that JARViS trained for a long-term VAD task is denoted as JARViS-LT. When 3D-CNN backbones are used, both JARViS and JARViS-LT outperform other VAD methods by significant margins. Note that JARViS-LT surpasses the current best method STMixer \cite{wu2023stmixer} by 2.4\% and 3.0\% on the AVA v2.1 and AVA v2.2, respectively.  When ViT backbones are used, the performance gains achieved by the proposed JARViS are maintained. JARViS model based on the ViT-B backbones yields 3.4\% higher mAP compared to STMixer on the AVA v2.2.


\subsubsection{Results on UCF101-24 and JHMDB51-21 datasets}
 Table \ref{table_ucf} and \ref{table_jhmdb} provide the performance of JARViS evaluated on UCF101-24 and JHMDB51-21 datasets, respectively. On UCF101-24, JARViS achieves 0.3\% and 0.9\% performance gains over ACAR and STMixer, respectively. Compared to ACAR, the performance gain of JARViS is relatively small. This result is likely due to the fact that ACAR used YOWO \cite{kopuklu2019you} as a person detector, which outperforms Faster-RCNN.
In contrast, on JHMDB51-21 dataset, JARViS achieves the SOTA performance with an impressive 2.4\% improvement in frame level mAP over the previous SOTA performance of STMixer. Furthermore, JARViS outperforms TubeR by 4.0\% and 6.9\% in video level mAP at IoU threshold of 0.2 and 0.5, respectively.

\subsection{Performance Analysis}
We used Faster R-CNN with a ResNeXt-101-FPN for a person detector and SlowFast-R50 for a video backbone in conducting the ablation studies. Models were trained and evaluated on AVA v2.2 dataset. We disabled temporal data augmentation except for Table \ref{abl_c}.

\begin{table}[t]
\caption{Ablation on the main components of JARViS}
\label{ablation1_check}
\centering
\resizebox{.6\columnwidth}{!}{
                            \begin{tabular}{ccccc}
                            \toprule
                            \multicolumn{1}{c}{ Separated}                & Unified     & Dense      &  Long-term & \multirow{2}{*}{mAP}  \\
                                                            feature spaces & transformer & proposals & context    &                       \\
                            \midrule
                                      &           &          & & 26.7   \\
                                      & \ding{51} &          &    & 27.0          \\
                                      & \ding{51} & \ding{51} &    & 27.0          \\
                            \ding{51} & \ding{51} &          &   & 28.5           \\
                            \ding{51} &           & \ding{51} &    & 29.5           \\
                            \ding{51} & \ding{51} & \ding{51} &    & 29.9    \\
                            \ding{51} & \ding{51} & \ding{51} & \ding{51}   & \textbf{31.2} \\
                            \bottomrule
                        \end{tabular}
}
\end{table}

\subsubsection{Ablation study}
    In our ablation study, we considered three components of JARViS: 1) network separation for actor features and scene context features, 2) unified Transformer, and 3) densely sampled actor proposals. Models with SlowFast-R50 backbone were evaluated on AVA v2.2 dataset. The proposed JARViS model achieves 29.9\% mAP.  Without all three components, VAD performance drops significantly to 26.7\%.   Table \ref{ablation1_check} presents the performance evaluated after removing each component from the original JARViS.  Without {\it Separated Feature Spaces}, VAD performance decreases by 2.9\%, which means that {\it Separated Feature Spaces} contributes crucially to overall performance.  In this case, we extracted both actor and scene context features from the shared video backbone features. Next, we replaced {\it unified Transformer} with a vanilla encoder-decoder Transformer \cite{vaswani2017attention}. The unified Transformer offers 0.3\% mAP gain over the baseline, which shows that the former is better at modeling interactions between two groups of embedding features. Finally, {\it densely sampled actor proposals} achieves 1.5\% performance gain over the case where the actor proposals are obtained from the final actor detection results of the person detector. When we train JARViS-LT, the performance improves by 1.3\%, which confirms the effectiveness of our weighted score aggregation strategy.

\begin{figure*}[t]

\centering

\begin{subfigure}[b]{0.3\textwidth}
\centering
\includegraphics[width=\textwidth]{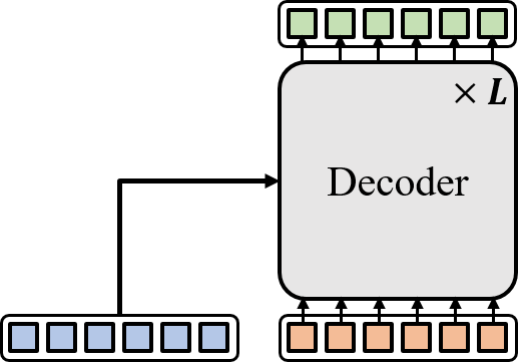}
\caption{Decoder only}
\label{fig:fig4a}
\end{subfigure}
\hfill
\begin{subfigure}[b]{0.3\textwidth}
\centering
\includegraphics[width=\textwidth]{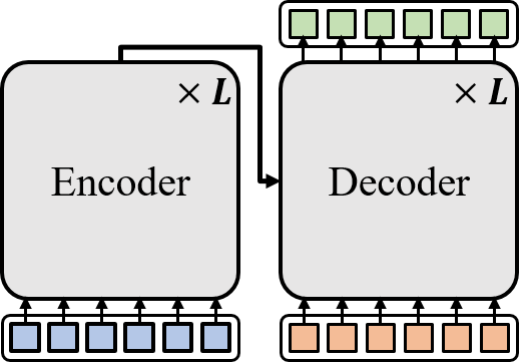}
\caption{Encoder-decoder}
\label{fig:fig4b}
\end{subfigure}
\hfill
\begin{subfigure}[b]{0.3\textwidth}
\centering
\includegraphics[width=\textwidth]{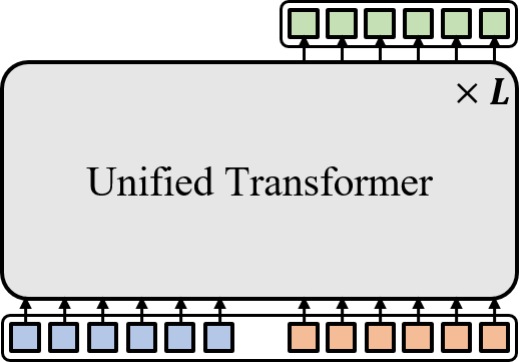}
\caption{Unified Transformer}
\label{fig:fig4c}
\end{subfigure}
\caption{Different relation modeling architectures.  Note that $L$ denotes the number of layers and the blue, red, and green tokens represent the scene context embedding, actor embedding, and their joint representations, respectively.}
\label{fig:fig4}
\end{figure*}

\begin{table}[t]
\caption{Comparison of different relation modeling architectures.}
\centering
\begin{adjustbox}{width=0.2\linewidth}
    \centering
    \begin{tabular}{@{} L{2cm} c @{}}
    \toprule
    Structure & mAP \\
    \midrule
    Dec   & 29.5 \\
    Enc-Dec & 29.5 \\
    Unified & \textbf{29.9} \\
    \bottomrule
    \end{tabular}
\end{adjustbox}
\label{abl_a}
\end{table}

\subsubsection{Relation modeling architectures}
    We compare our unified actor-scene context Transformer with several other relation modeling architectures. In Table \ref{abl_a}, we present the performances of 1) the {\it decoder only Transformer} (i.e., cross-attention only), 2) the vanilla {\it encoder-decoder Transformer}, and 3) the proposed {\it unified Transformer}.
    These three models are depicted in Fig. \ref{fig:fig4} and were used for relation modeling of JARViS. The unified Transformer used in JARViS offers 0.4\% mAP gain over both {\it decoder only Transformer} and {\it encoder-decoder Transformer}.

\begin{table}[t]
\caption{Ablation study of matching cost components.}
\centering
\begin{adjustbox}{width=0.6\linewidth}
    \centering
     \begin{tabular}{ccccccc}
    \toprule

    $\mathcal{L}_{\text{cls}}^\text{action}$ & $\mathcal{L}_{\text{cls}}^\text{person}$ & $\mathcal{L}_{\text{box}}$ &mAP     &mAP$_{\text{p}}$ &mAP$_{\text{pp}}$&mAP$_{\text{po}}$ \\
    \midrule

    \ding{51}   &             & \ding{51}    &  29.5           & 49.8   & 31.5    & 20.3   \\
                 & \ding{51}  & \ding{51}    &  \textbf{29.9}  &  \textbf{50.4}   &  \textbf{31.8}    &  \textbf{20.6}   \\
    \ding{51}   & \ding{51}  &  \ding{51}    &  29.7           & 50.1   & 31.7    & 20.4   \\

    \bottomrule
    \end{tabular}
\end{adjustbox}
\label{matching}
\end{table}

\subsubsection{Matching cost}
\label{sec:matching_cost}
In  Table \ref{matching}, we used the person detection loss, $\mathcal{L}_{\text{cls}}(h_i^{gt},\hat{h}_{j})$, and the bounding box regression loss $\mathcal{L}_{\text{box}}(b_i^{gt},\hat{b}_{j})$ for the matching cost.  We can construct a different matching cost by taking a different combination from the action classification loss $\mathcal{L}_{\text{cls}}^{\rm action}=\mathcal{L}_{\text{cls}}(c_i^{gt},\hat{c}_{j})$, the person detection loss, $\mathcal{L}_{\text{cls}}^{\rm person}=\mathcal{L}_{\text{cls}}(h_i^{gt},\hat{h}_{j})$, and the bounding box regression loss $\mathcal{L}_{\text{box}}=\mathcal{L}_{\text{box}}(b_i^{gt},\hat{b}_{j})$.
Table \ref{matching} compares the performance of JARViS when different combinations are used for the matching cost. We considered the performances on three action subcategories \cite{gu2018ava}, i.e., {\it person poses}, {\it person-person interactions}, and {\it person-object interactions}. The mAPs evaluated on these three subcategories are denoted as mAP$_{\text{p}}$, mAP$_{\text{pp}}$, and mAP$_{\text{po}}$. We see that  using  $\mathcal{L}_{\text{cls}}^{\rm person}$ and $\mathcal{L}_{\text{box}}$ yields better performance than using $\mathcal{L}_{\text{cls}}^{\rm action}$ and $\mathcal{L}_{\text{box}}$. Using all three loss functions does not promise the best performance. These results demonstrate that our design is best able to identify correct permutations of action proposals.

\begin{table}[t]
\caption{Effect of the number of Transformer layers.}
\centering
\begin{adjustbox}{width=0.4\linewidth}
\centering
    \begin{tabular}{@{} c c c c @{}}
    \toprule
    \#layers & \#params & GFLOPs & mAP \\
    \midrule
    1 & 139.3M & 206.7 & 28.7 \\
    3 & 140.9M & 210.4 & 28.9 \\
    6 & 143.3M & 216.1& \textbf{29.9} \\
    12 & 148.0M & 227.3& \textbf{29.9} \\
    \bottomrule
    \end{tabular}
\end{adjustbox}
\label{abl_b}
\end{table}

\subsubsection{Transformer layer depth}
      Table \ref{abl_b} presents the performance of JARViS as a function of the number of Transformer layers. We report the mean GFLOPs for the initial 100 video clips in the AVA validation set.
      Performance improves as the number of transformer layers increases, but the performance improvement diminishes as the number of layers exceeds 6.

\begin{table}[t!]
\caption{Ablation study of proposal sampling strategies.}
\centering
\begin{adjustbox}{width=0.7\linewidth}
\centering

             \begin{tabular}{lcccccc}
    \toprule

    Confidence threshold                       & 0.9      &   0.8      &  0.7    &   0.5     &   0.1    &  0 (top-K)    \\

    \midrule
    mAP                                        &  28.6    &  29.3      &  29.6   &   29.7    &   29.8   &   \textbf{29.9}    \\

    \bottomrule

            \end{tabular}
\end{adjustbox}

\label{dense_sampling}
\end{table}

\subsubsection{Densely sampled proposals}
To evaluate the effectiveness of using densely sampled actor proposals, we varied the density of actor proposals. To this end, we adjusted the confidence threshold of the person detector.
As the confidence threshold decreases, the number of actor proposal features generated by the person detector increases. Table \ref{dense_sampling} presents the performance of JARViS for several values of confidence threshold.
Notably, the best performance was observed when utilizing a dense sampling strategy without applying a confidence threshold. This demonstrates JARViS's ability to effectively distinguish between positive and negative samples within the densely sampled actor proposals.

\subsubsection{Duration of temporal support}
Table \ref{longterm} shows the benefits of incorporating long-term contexts through weighted score aggregation. The results are presented for various duration of temporal support. We set a baseline with a short-term video clip duration of 2.1 seconds and then experimented with increasing temporal support, ranging from 4 seconds to 16 seconds. We observed the best performance of 31.2\% mAP when using temporal support of 12 seconds. We note that the performance remained consistent even with further an extension of the temporal support duration.

\begin{table}[t!]
\caption{Ablation study of temporal support duration.}
\centering
\begin{adjustbox}{width=0.7\linewidth}
\centering

             \begin{tabular}{lcccccc}
    \toprule

    Temporal support (sec.)         &2     & 4    &6   & 8     & 12      &16     \\

    \midrule
    mAP                             &29.9  & 30.9 &  31.0  & 31.1  & \textbf{31.2}   &  \textbf{31.2} \\

    \bottomrule

            \end{tabular}
\end{adjustbox}
\label{longterm}
\end{table}

\subsubsection{Weighted score aggregation strategies}
We tried several score aggregation strategies for {\it weighted score aggregation method} for long-term VAD. We compare four operations, including 1) max pooling, 2) average pooling, 3) top-k pooling, and 4) weighted sum. Table \ref{abl_c} confirms that the weighted sum operation used for JARViS yields the best VAD performance.

\begin{table}[t!]
\caption{Score aggregation strategies.}
\centering
\begin{adjustbox}{width=0.35\linewidth}
    \centering
    \begin{tabular}{@{} L{4cm} c @{}}
    \toprule
    Aggregation & mAP \\
    \midrule
    Max & 25.1 \\
    Avg & 27.4 \\
    top-K pooling & 28.1 \\
    Weighted sum & \textbf{31.2} \\
    \bottomrule
    \end{tabular}
\end{adjustbox}
\label{abl_c}
\end{table}

\begin{table}[t!]
\caption{Offset ranges for temporal data augmentation.}
\centering
\begin{adjustbox}{width=0.3\linewidth}
    \centering
    \begin{tabular}{@{} C{3cm} c @{}}
    \toprule
    Range of $\Delta$ & mAP \\
    \midrule
    \text{[-1, 1]} & 29.9 \\
    \text{[-1.5, 1.5]} & \textbf{30.2} \\
    \text{[-2, 2]} & 29.8 \\
    \text{[-2.5, 2.5]} & 29.6 \\
    \bottomrule
    \end{tabular}
\end{adjustbox}
\label{abl_d}
\end{table}

\subsubsection{Temporal data augmentation}
Table \ref{abl_d} shows the performance of JARViS with different offset ranges for temporal data augmentation. We see that JARViS achieves the best performance when the offset range is set to $[-1.5,1.5]$.


\subsubsection{Computational efficiency}
We compare the computational efficiency of our JARViS with previous SOTA VAD methods in Table \ref{gflops}.
The total computational costs of the two-stage methods are denoted as the sum of the FLOPs of the action classifiers with the additional person detectors.
For a fair comparison, we maintain the resolution of square input frames (e.g., $224^2$, $256^2$) used to obtain the inference performance of each model and excluded VAD methods with long-term context.
The computational costs of conventional two-stage methods were calculated without the person detector component due to the lack of detailed implementation information.
One-stage VAD methods such as TubeR and STMixer demonstrated higher efficiency compared to conventional two-stage methods, considering the computational resources of F-RCNN. However, adopting a modern lightweight person detector empowers the two-stage approach to achieve competitive person detection performance while reducing computational costs. In Table \ref{det_compare}, DETR's person detection performance is slightly 0.4\% lower than F-RCNN's, yet it still outperforms end-to-end VAD models.

\begin{table}[t]
\caption{Performance comparison in terms of computational costs on AVA dataset. STMixer$^\dagger$ uses the resolution of the input frames from the ViT-B to $256^2$ instead of $224^2$.}
\centering
\begin{adjustbox}{width=0.6\linewidth}
\begin{tabular}{lccccc}
\toprule
\multicolumn{1}{c}{Model} & Detector & Input & Backbone   & GFLOPs   & mAP  \\
\midrule
SlowFast                  & F-RCNN   & $32 \times 2$  & SF-R101-NL & 119 + NA      & 29.0 \\
ACAR                      & F-RCNN   & $32 \times 2$  & SF-R101    & 160 + NA      & 31.7 \\
VideoMAE                  & F-RCNN   & $16 \times 4$  & ViT-B      & 180 + NA      & 31.8 \\
WOO                       & \xmark        & $32 \times 2$  & SF-R101-NL & 252      & 28.3 \\
TubeR                     & \xmark        & $32 \times 2$  & CSN-152    & 120      & 31.1 \\
STMixer                   & \xmark        & $32 \times 2$  & SF-R101    & 135      & 30.1 \\
STMixer$^\dagger$            & \xmark        & $16 \times 4$  & $\text{ViT-B}^*$        & 355      & 36.1 \\
\midrule
JARViS                    & F-RCNN   & $32 \times 2$  & SF-R101    & 166 + 67 & 35.1 \\
JARViS                    & F-RCNN   & $16 \times 4$  & ViT-B      & 196 + 67 & 35.4\\
JARViS                    & DETR     & $32 \times 2$  & SF-R101    & 166 + 12 & 34.9 \\
JARViS                    & DETR     & $16 \times 4$  & $\text{ViT-B}^*$      & 196 + 12 & \textbf{39.5}\\
\bottomrule
\end{tabular}
\end{adjustbox}

\label{gflops}
\end{table}

\begin{table}[t]
\caption{Person detection performance of end-to-end VAD models and pre-trained person detectors on AVA dataset.}
\centering
\begin{adjustbox}{width=0.5\linewidth}
\begin{tabular}{lccccc}
\toprule
\multicolumn{1}{c}{Model} & Backbone                       & AP (person)      \\

\midrule
WOO                    & SF-R101                           & 95.6      \\
TubeR                  & CSN-152                           & 93.8      \\
STMixer                & $\text{ViT-B}^*$                  & 94.1       \\
F-RCNN                 & ResNeXt-101-FPN                   & \textbf{96.7}      \\
DETR                   & ResNet-50                         & 96.3       \\
\bottomrule
\end{tabular}
\end{adjustbox}

\label{det_compare}
\end{table}

\begin{figure*}[t]
\centering
\includegraphics[width=0.99\textwidth]{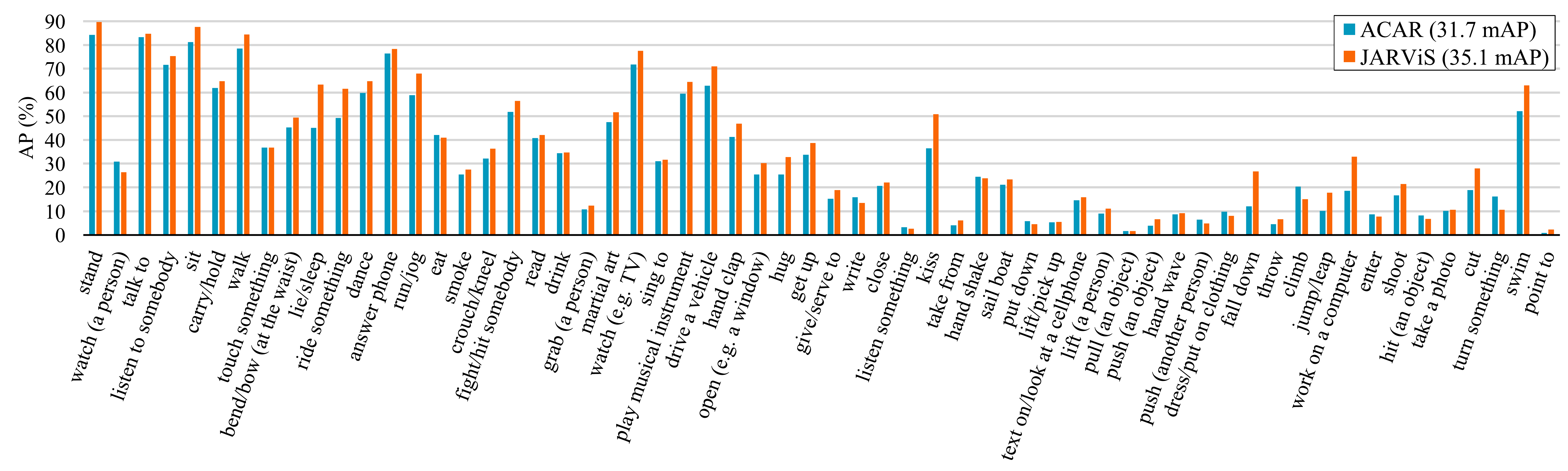}
\caption{Performance of JARViS versus ACAR for each action class on the AVA v2.2.}
\label{acar_vs}
\end{figure*}

\begin{figure*}[t]
\centering
\includegraphics[width=0.99\textwidth]{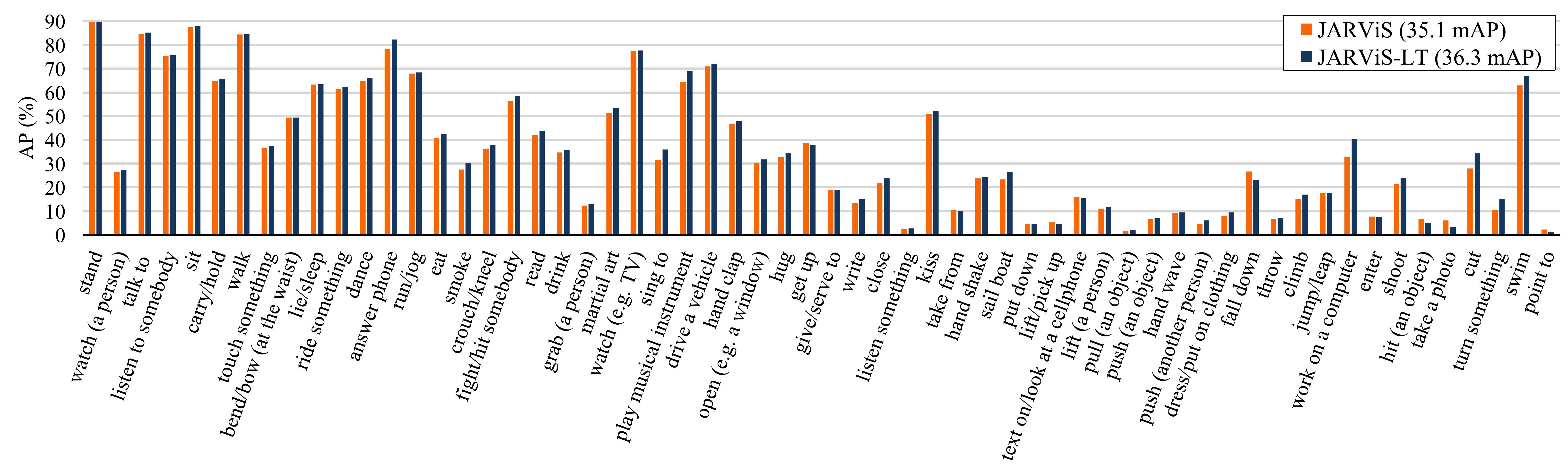}
\caption{Impact of long-term score aggregation on the AVA v2.2.}
\label{lt_vs}
\end{figure*}

\begin{figure}[t]
\centering
\includegraphics[width=0.9\textwidth]{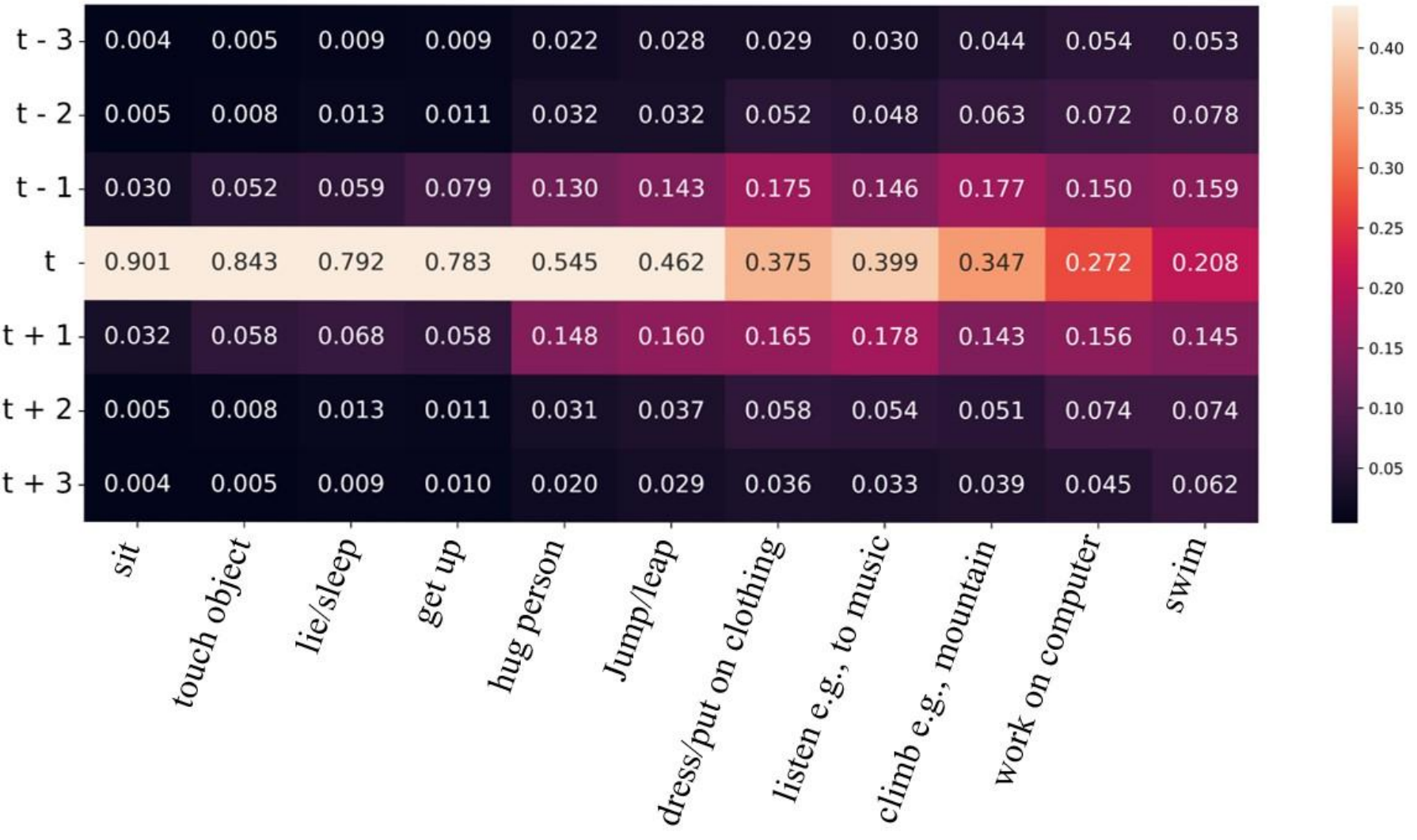}
\caption{Heat map of score aggregation weights. The brightness of each box reflects its weight value, which means that a brighter color indicates a higher weight.}
\label{score}
\end{figure}

\subsubsection{Weighted score aggregation}
Fig. \ref{score} provides the values of the optimized score aggregation weights for different time steps and action classes in JARViS-LT. The keyframe is located at time step $t$, so the score aggregation weight has the largest value at time step $t$. Note that the score aggregation weights have different distributions in class. For momentary actions such as `touching objects' or `jumping', the score aggregation weights have significantly large values around time step $t$. For dynamic actions that span over a longer period of time, such as `swimming' or `climbing mountain', past and future scores were broadly taken into account.

\subsubsection{Per-class comparison on AVA dataset}
We compare our JARViS with ACAR \cite{pan2021actor}  in the per-class AP on AVA v2.2 dataset. For a fair comparison, the SF-R101 backbone network was used for both models. Fig. \ref{acar_vs} shows that JARViS outperforms ACAR in many classes. In particular, within the {\it person poses} category, significant performance gains are observed in classes such as `sleep', `fall down', and `jump'. In the {\it person-person interactions}, JARViS achieves higher performance in `kiss' and `hug', where visual occlusion can occur. Notably, considerable performance gains are observed in  {\it person-object interactions}, particularly in challenging classes (e.g., `cut' and `work on a computer').
Fig. \ref{lt_vs} compares per-class performance between short-term and long-term prediction results of JARViS. Utilizing long-term video clips improves overall mAP performance by 1.2\%. The performance of JARViS-LT improves in 51 out of 60 classes compared to its short-term version. The most significant improvements are observed in the following five classes, `work on a computer' (+7.42\%), `turn something' (+4.55\%), `swim' (+4.47\%), `sing to' (+4.27\%), and `answer phone' (+3.86\%).

\subsection{Visual results}
\begin{figure*}[t]
\centering
\includegraphics[width=0.99\textwidth]{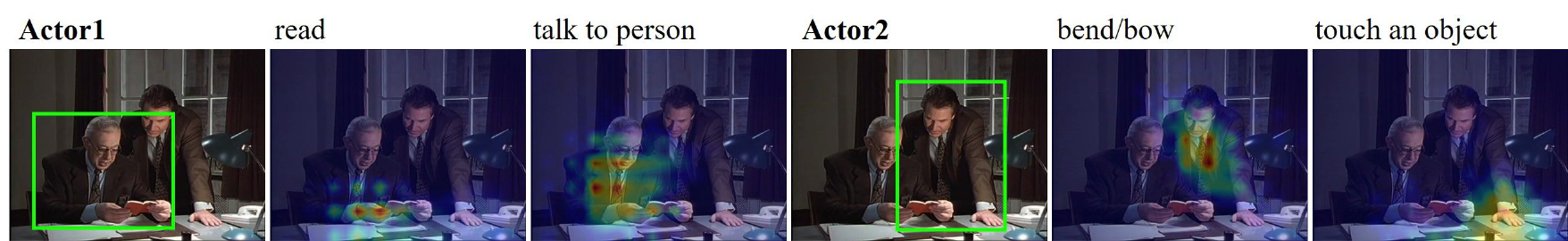}
\caption{Visualization of activation maps. The activation maps generated for some action classes are visualized on a few keyframes of AVA v2.2.  The activated regions are highly relevant to each action label.
}
\label{fig:fig5}
\end{figure*}

\begin{figure}[!h]
\centering
\includegraphics[width=0.9\textwidth]{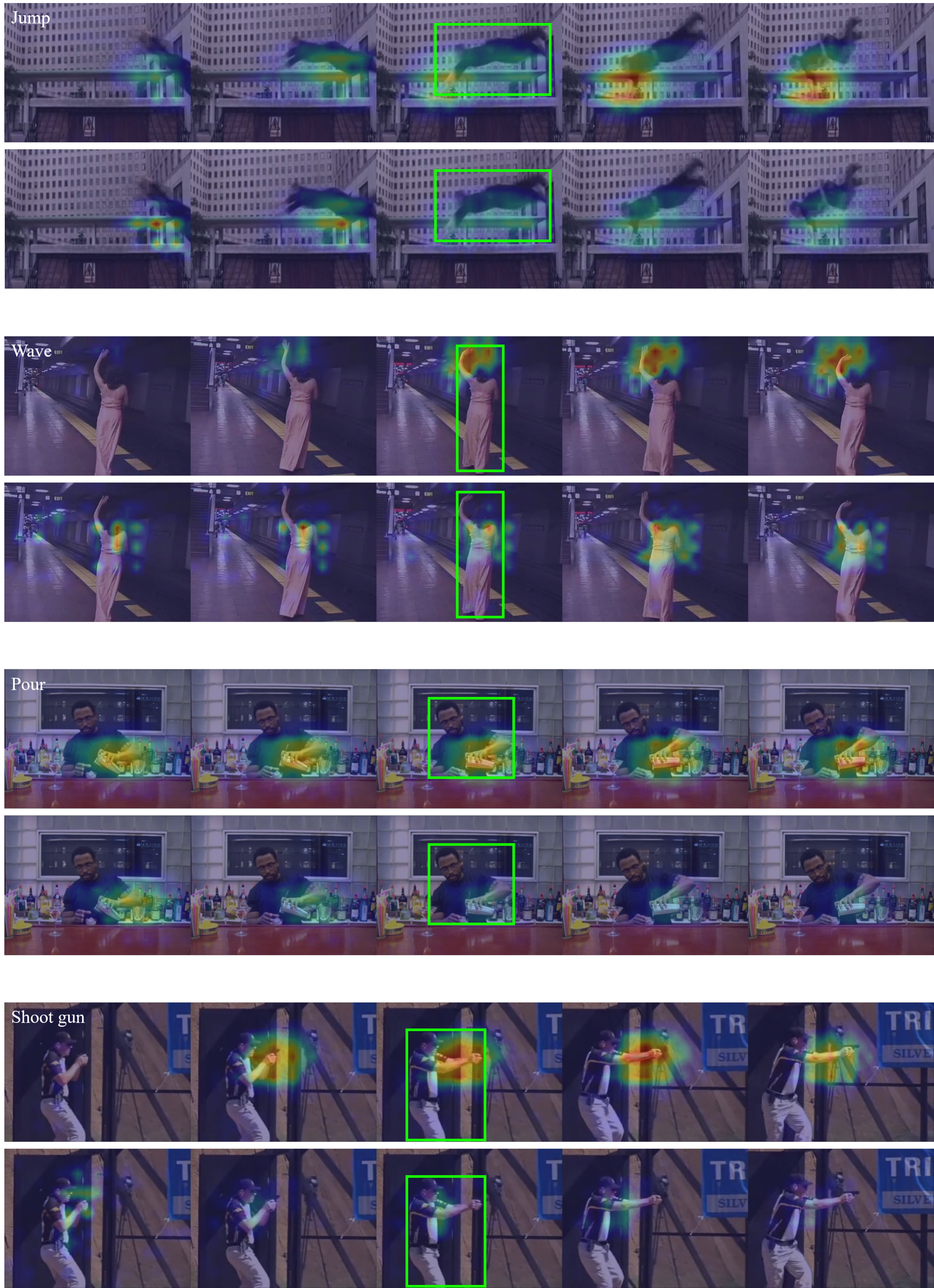}
\caption{Visualization of activation maps per frame. The activation maps generated for some action classes are visualized on video clips of JHMDB51-21.}
\label{heatmap_vis}
\end{figure}

In Fig. \ref{fig:fig5}, we visualize the class attention maps \cite{zhou2016learning} generated by a unified actor-scene context transformer. We see that JARViS captures discriminated regions to generate the action proposal features. As depicted in Fig. \ref{fig:fig5}, JARViS concentrates on the relational modeling between person and object or person and person, in classes like 'read' and 'touch an object' which belong to person-object or person-person categories. Furthermore, for classes such as 'stand' and 'bend' which are part of the person poses category, it is observed that the activation map is generated with a focus on the actor performing the action.
Fig. \ref{heatmap_vis} also shows visualizations of the attention maps, where examples were chosen from the JHMDB51-21 dataset. The top row of the visualization is the result of JARViS, while the bottom row shows the results of the conventional two-stage methods, which sample RoI features from the shared backbone network. The results show that leveraging independent actor and scene semantics is much better at capturing action cues in space and time.

\begin{figure}[!h]
\centering
\includegraphics[width=0.7\textwidth]{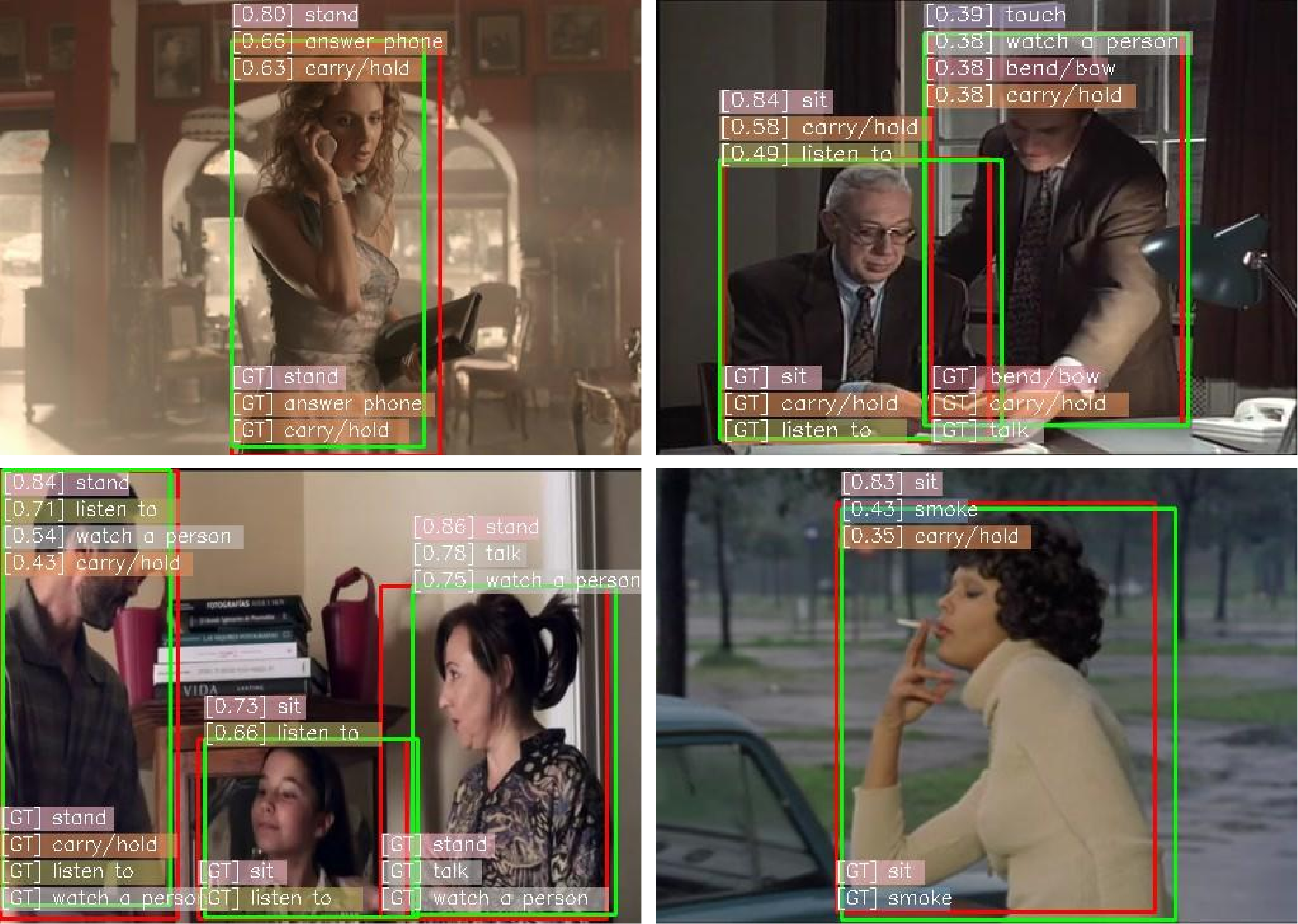}
\caption{Visualization of detection results on AVA. The ground truth and prediction boxes for the actors' actions are outlined in red and green, respectively. The predicted classes for each actor are shown above the box.}
\label{results_vis}
\end{figure}

In Fig. \ref{results_vis}, we provide the visualization of the detection results of JARViS. Note that JARViS can even detect an action that was missing from AVA annotations in the bottom-right figure, where a smoking woman is holding a cigarette.

\section{Conclusions}
In this paper, we proposed a novel two-stage VAD framework, JARViS, which can effectively utilize scene contexts to achieve fine-grained action detection. JARViS independently generates actor features and scene context features using different backbone networks and reasons about their inter-dependencies through a unified Transformer structure. The unified Transformer infers the fine-grained relationships between densely sampled actor features and global-scale scene features. Utilizing the bipartite matching loss, it then generates a finite set of action predictions.
In addition, we introduced a weighted score aggregation method to gather long-term contextual information from extended video clips. Our experiments on popular VAD benchmarks demonstrated that our key ideas have notably boosted baseline performance, with JARViS attaining state-of-the-art results over current VAD methods.

The framework proposed in JARViS opens up the possibility of extending the relational modeling framework with additional features and input sources, which can take VAD to the next level of performance. In the extended studies, visual features such as object-related semantics or pose-related semantics (e.g., joint skeletons) and auditory semantics can be added to enhance relation modeling for VAD.

\bibliographystyle{elsarticle-num}
\bibliography{ref}

\begin{thebibliography}{10}
\expandafter\ifx\csname url\endcsname\relax
  \def\url#1{\texttt{#1}}\fi
\expandafter\ifx\csname urlprefix\endcsname\relax\def\urlprefix{URL }\fi
\expandafter\ifx\csname href\endcsname\relax
  \def\href#1#2{#2} \def\path#1{#1}\fi

\bibitem{vahdani2022deep}
E.~Vahdani, Y.~Tian, Deep learning-based action detection in untrimmed videos: a survey, IEEE Transactions on Pattern Analysis and Machine Intelligence (2022).

\bibitem{carreira2017quo}
J.~Carreira, A.~Zisserman, Quo vadis, action recognition? a new model and the kinetics dataset, in: proceedings of the IEEE Conference on Computer Vision and Pattern Recognition, 2017, pp. 6299--6308.

\bibitem{feichtenhofer2019slowfast}
C.~Feichtenhofer, H.~Fan, J.~Malik, K.~He, Slowfast networks for video recognition, in: Proceedings of the IEEE/CVF International Conference on Computer Vision, 2019, pp. 6202--6211.

\bibitem{tong2022videomae}
Z.~Tong, Y.~Song, J.~Wang, L.~Wang, Videomae: Masked autoencoders are data-efficient learners for self-supervised video pre-training, Advances in neural information processing systems 35 (2022) 10078--10093.

\bibitem{sun2018actor}
C.~Sun, A.~Shrivastava, C.~Vondrick, K.~Murphy, R.~Sukthankar, C.~Schmid, Actor-centric relation network, in: Proceedings of the European Conference on Computer Vision (ECCV), 2018, pp. 318--334.

\bibitem{chen2021watch}
S.~Chen, P.~Sun, E.~Xie, C.~Ge, J.~Wu, L.~Ma, J.~Shen, P.~Luo, Watch only once: An end-to-end video action detection framework, in: Proceedings of the IEEE/CVF International Conference on Computer Vision, 2021, pp. 8178--8187.

\bibitem{zhao2022tuber}
J.~Zhao, Y.~Zhang, X.~Li, H.~Chen, B.~Shuai, M.~Xu, C.~Liu, K.~Kundu, Y.~Xiong, D.~Modolo, et~al., Tuber: Tubelet transformer for video action detection, in: Proceedings of the IEEE/CVF Conference on Computer Vision and Pattern Recognition, 2022, pp. 13598--13607.

\bibitem{wu2023stmixer}
T.~Wu, M.~Cao, Z.~Gao, G.~Wu, L.~Wang, Stmixer: A one-stage sparse action detector, in: Proceedings of the IEEE/CVF Conference on Computer Vision and Pattern Recognition, 2023, pp. 14720--14729.

\bibitem{girdhar2019video}
R.~Girdhar, J.~Carreira, C.~Doersch, A.~Zisserman, Video action transformer network, in: Proceedings of the IEEE/CVF Conference on Computer Vision and Pattern Recognition, 2019, pp. 244--253.

\bibitem{wu2019long}
C.-Y. Wu, C.~Feichtenhofer, H.~Fan, K.~He, P.~Krahenbuhl, R.~Girshick, Long-term feature banks for detailed video understanding, in: Proceedings of the IEEE/CVF Conference on Computer Vision and Pattern Recognition, 2019, pp. 284--293.

\bibitem{pan2021actor}
J.~Pan, S.~Chen, M.~Z. Shou, Y.~Liu, J.~Shao, H.~Li, Actor-context-actor relation network for spatio-temporal action localization, in: Proceedings of the IEEE/CVF Conference on Computer Vision and Pattern Recognition, 2021, pp. 464--474.

\bibitem{vaswani2017attention}
A.~Vaswani, N.~Shazeer, N.~Parmar, J.~Uszkoreit, L.~Jones, A.~N. Gomez, {\L}.~Kaiser, I.~Polosukhin, Attention is all you need, Advances in Neural Information Processing Systems 30 (2017).

\bibitem{dosovitskiy2021an}
A.~Dosovitskiy, L.~Beyer, A.~Kolesnikov, D.~Weissenborn, X.~Zhai, T.~Unterthiner, M.~Dehghani, M.~Minderer, G.~Heigold, S.~Gelly, J.~Uszkoreit, N.~Houlsby, An image is worth 16x16 words: Transformers for image recognition at scale, in: International Conference on Learning Representations, 2021.

\bibitem{zhou2020unified}
L.~Zhou, H.~Palangi, L.~Zhang, H.~Hu, J.~Corso, J.~Gao, {Unified vision-language pre-training for image captioning and VQA}, in: Proceedings of the AAAI conference on artificial intelligence, Vol.~34, 2020, pp. 13041--13049.

\bibitem{carion2020end}
N.~Carion, F.~Massa, G.~Synnaeve, N.~Usunier, A.~Kirillov, S.~Zagoruyko, End-to-end object detection with transformers, in: Computer Vision--ECCV 2020: 16th European Conference, Glasgow, UK, August 23--28, 2020, Proceedings, Part I 16, Springer, 2020, pp. 213--229.

\bibitem{li2020actions}
Y.~Li, Z.~Wang, L.~Wang, G.~Wu, Actions as moving points, in: Computer Vision--ECCV 2020: 16th European Conference, Glasgow, UK, August 23--28, 2020, Proceedings, Part XVI 16, Springer, 2020, pp. 68--84.

\bibitem{gu2018ava}
C.~Gu, C.~Sun, D.~A. Ross, C.~Vondrick, C.~Pantofaru, Y.~Li, S.~Vijayanarasimhan, G.~Toderici, S.~Ricco, R.~Sukthankar, et~al., Ava: A video dataset of spatio-temporally localized atomic visual actions, in: Proceedings of the IEEE Conference on Computer Vision and Pattern Recognition, 2018, pp. 6047--6056.

\bibitem{zhang2019structured}
Y.~Zhang, P.~Tokmakov, M.~Hebert, C.~Schmid, A structured model for action detection, in: Proceedings of the IEEE/CVF Conference on Computer Vision and Pattern Recognition, 2019, pp. 9975--9984.

\bibitem{tang2020asynchronous}
J.~Tang, J.~Xia, X.~Mu, B.~Pang, C.~Lu, Asynchronous interaction aggregation for action detection, in: Computer Vision--ECCV 2020: 16th European Conference, Glasgow, UK, August 23--28, 2020, Proceedings, Part XV 16, Springer, 2020, pp. 71--87.

\bibitem{chen2023efficient}
L.~Chen, Z.~Tong, Y.~Song, G.~Wu, L.~Wang, Efficient video action detection with token dropout and context refinement, in: Proceedings of the IEEE/CVF International Conference on Computer Vision, 2023, pp. 10388--10399.

\bibitem{gritsenko2024end}
A.~A. Gritsenko, X.~Xiong, J.~Djolonga, M.~Dehghani, C.~Sun, M.~Lucic, C.~Schmid, A.~Arnab, End-to-end spatio-temporal action localisation with video transformers, in: Proceedings of the IEEE/CVF Conference on Computer Vision and Pattern Recognition, 2024, pp. 18373--18383.

\bibitem{wang2023videomae}
L.~Wang, B.~Huang, Z.~Zhao, Z.~Tong, Y.~He, Y.~Wang, Y.~Wang, Y.~Qiao, Videomae v2: Scaling video masked autoencoders with dual masking, in: Proceedings of the IEEE/CVF Conference on Computer Vision and Pattern Recognition, 2023, pp. 14549--14560.

\bibitem{wang2023masked}
R.~Wang, D.~Chen, Z.~Wu, Y.~Chen, X.~Dai, M.~Liu, L.~Yuan, Y.-G. Jiang, Masked video distillation: Rethinking masked feature modeling for self-supervised video representation learning, in: Proceedings of the IEEE/CVF Conference on Computer Vision and Pattern Recognition, 2023, pp. 6312--6322.

\bibitem{ryali2023hiera}
C.~Ryali, Y.-T. Hu, D.~Bolya, C.~Wei, H.~Fan, P.-Y. Huang, V.~Aggarwal, A.~Chowdhury, O.~Poursaeed, J.~Hoffman, et~al., Hiera: A hierarchical vision transformer without the bells-and-whistles, in: International Conference on Machine Learning, PMLR, 2023, pp. 29441--29454.

\bibitem{he2022masked}
K.~He, X.~Chen, S.~Xie, Y.~Li, P.~Doll{\'a}r, R.~Girshick, Masked autoencoders are scalable vision learners, in: Proceedings of the IEEE/CVF Conference on Computer Vision and Pattern Recognition, 2022, pp. 16000--16009.

\bibitem{he2017mask}
K.~He, G.~Gkioxari, P.~Doll{\'a}r, R.~Girshick, Mask r-cnn, in: Proceedings of the IEEE International Conference on Computer Vision, 2017, pp. 2961--2969.

\bibitem{ren2015faster}
S.~Ren, K.~He, R.~Girshick, J.~Sun, Faster r-cnn: Towards real-time object detection with region proposal networks, Advances in Neural Information Processing Systems 28 (2015).

\bibitem{kay2017kinetics}
W.~Kay, J.~Carreira, K.~Simonyan, B.~Zhang, C.~Hillier, S.~Vijayanarasimhan, F.~Viola, T.~Green, T.~Back, P.~Natsev, et~al., The kinetics human action video dataset, arXiv preprint arXiv:1705.06950 (2017).

\bibitem{carreira2019short}
J.~Carreira, E.~Noland, C.~Hillier, A.~Zisserman, A short note on the kinetics-700 human action dataset, arXiv preprint arXiv:1907.06987 (2019).

\bibitem{ba2016layer}
J.~L. Ba, J.~R. Kiros, G.~E. Hinton, Layer normalization, arXiv preprint arXiv:1607.06450 (2016).

\bibitem{zhu2021deformable}
X.~Zhu, W.~Su, L.~Lu, B.~Li, X.~Wang, J.~Dai, Deformable detr: Deformable transformers for end-to-end object detection, in: International Conference on Learning Representations, 2021.

\bibitem{lin2017focal}
T.-Y. Lin, P.~Goyal, R.~Girshick, K.~He, P.~Doll{\'a}r, Focal loss for dense object detection, in: Proceedings of the IEEE International Conference on Computer Vision, 2017, pp. 2980--2988.

\bibitem{rezatofighi2019generalized}
H.~Rezatofighi, N.~Tsoi, J.~Gwak, A.~Sadeghian, I.~Reid, S.~Savarese, Generalized intersection over union: A metric and a loss for bounding box regression, in: Proceedings of the IEEE/CVF Conference on Computer Vision and Pattern Recognition, 2019, pp. 658--666.

\bibitem{kuhn1955hungarian}
H.~W. Kuhn, The hungarian method for the assignment problem, Naval research logistics quarterly 2~(1-2) (1955) 83--97.

\bibitem{soomro2012ucf101}
K.~Soomro, A.~R. Zamir, M.~Shah, {UCF101: A dataset of 101 human actions classes from videos in the wild}, arXiv preprint arXiv:1212.0402 (2012).

\bibitem{jhuang2013towards}
H.~Jhuang, J.~Gall, S.~Zuffi, C.~Schmid, M.~J. Black, Towards understanding action recognition, in: Proceedings of the IEEE International Conference on Computer Vision, 2013, pp. 3192--3199.

\bibitem{lin2017feature}
T.-Y. Lin, P.~Doll{\'a}r, R.~Girshick, K.~He, B.~Hariharan, S.~Belongie, Feature pyramid networks for object detection, in: Proceedings of the IEEE Conference on Computer Vision and Pattern Recognition, 2017, pp. 2117--2125.

\bibitem{xie2017aggregated}
S.~Xie, R.~Girshick, P.~Doll{\'a}r, Z.~Tu, K.~He, Aggregated residual transformations for deep neural networks, in: Proceedings of the IEEE Conference on Computer Vision and Pattern Recognition, 2017, pp. 1492--1500.

\bibitem{girshick2018detectron}
R.~Girshick, I.~Radosavovic, G.~Gkioxari, P.~Doll{\'a}r, K.~He, Detectron (2018).

\bibitem{he2016deep}
K.~He, X.~Zhang, S.~Ren, J.~Sun, Deep residual learning for image recognition, in: Proceedings of the IEEE conference on computer vision and pattern recognition, 2016, pp. 770--778.

\bibitem{russakovsky2015imagenet}
O.~Russakovsky, J.~Deng, H.~Su, J.~Krause, S.~Satheesh, S.~Ma, Z.~Huang, A.~Karpathy, A.~Khosla, M.~Bernstein, et~al., Imagenet large scale visual recognition challenge, International journal of computer vision 115 (2015) 211--252.

\bibitem{lin2014microsoft}
T.-Y. Lin, M.~Maire, S.~Belongie, J.~Hays, P.~Perona, D.~Ramanan, P.~Doll{\'a}r, C.~L. Zitnick, Microsoft coco: Common objects in context, in: Computer Vision--ECCV 2014: 13th European Conference, Zurich, Switzerland, September 6-12, 2014, Proceedings, Part V 13, Springer, 2014, pp. 740--755.

\bibitem{loshchilov2017decoupled}
I.~Loshchilov, F.~Hutter, Decoupled weight decay regularization, arXiv preprint arXiv:1711.05101 (2017).

\bibitem{kalogeiton2017action}
V.~Kalogeiton, P.~Weinzaepfel, V.~Ferrari, C.~Schmid, Action tubelet detector for spatio-temporal action localization, in: Proceedings of the IEEE International Conference on Computer Vision, 2017, pp. 4405--4413.

\bibitem{yang2019step}
X.~Yang, X.~Yang, M.-Y. Liu, F.~Xiao, L.~S. Davis, J.~Kautz, Step: Spatio-temporal progressive learning for video action detection, in: Proceedings of the IEEE/CVF Conference on Computer Vision and Pattern Recognition, 2019, pp. 264--272.

\bibitem{kopuklu2019you}
O.~K{\"o}p{\"u}kl{\"u}, X.~Wei, G.~Rigoll, You only watch once: A unified cnn architecture for real-time spatiotemporal action localization, arXiv preprint arXiv:1911.06644 (2019).

\bibitem{zhou2016learning}
B.~Zhou, A.~Khosla, A.~Lapedriza, A.~Oliva, A.~Torralba, Learning deep features for discriminative localization, in: Proceedings of the IEEE conference on computer vision and pattern recognition, 2016, pp. 2921--2929.

\end{thebibliography}





\end{document}